\documentclass{article}

\usepackage[final]{neurips_data_2022}

\usepackage[utf8]{inputenc} 
\usepackage[T1]{fontenc}    
\usepackage{hyperref}       
\usepackage{url}            
\usepackage{booktabs}       
\usepackage{amsfonts}       
\usepackage{nicefrac}       
\usepackage{microtype}      
\usepackage{xcolor}         
\usepackage{makecell}
\usepackage{graphicx}
\usepackage[T1]{tipa}

\usepackage{substitutefont}

\substitutefont{T3}{lmr}{cmr}
\graphicspath{ {./images/} }

\newcommand{\benchmarkname}{LEPISZCZE }
\newcommand{\clarinplbiz}{CLARIN-PL-Biz }
\newcommand{\clarinpl}{CLARIN-PL }
\newcommand{\libraryname}{\texttt{clarinpl-embeddings} }
\newcommand{\repositoryurl}{ https://github.com/CLARIN-PL/LEPISZCZE }
\newcommand{\weightbiasesurl}{https://wandb.ai/embeddings/LEPISZCZE }

\newcommand{\mathBF}[1]{%
    \pdfliteral direct {2 Tr 0.3 w} 
     #1%
    \pdfliteral direct {0 Tr 0 w}%
}

\title{This is the way: designing and compiling LEPISZCZE, a comprehensive NLP benchmark for Polish}

\author{%
  Łukasz Augustyniak \\
  WUST \\ (Wroclaw University of \\ Science and Technology)\\
   \And
   Kamil Tagowski \\
   WUST\\
   \And
   Albert Sawczyn \\
   WUST\\
   \And
   Denis Janiak \\
   WUST\\
   \And
   Roman Bartusiak \\
   WUST\\
   \And
   Adrian Szymczak \\
   WUST\\
   \And
   Marcin Wątroba \\
   WUST\\
   \And
   Arkadiusz Janz \\
   WUST\\
   \And
    Piotr Szymański \\
   WUST\\
   \And
   Mikołaj Morzy \\
   Poznan University of Technology\\
    \And
   Tomasz Kajdanowicz \\
   WUST\\
\And
   Maciej Piasecki \\
   WUST\\
}

\begin{document}
\maketitle
\begin{abstract}

The availability of compute and data to train larger and larger language models increases the demand for robust methods of benchmarking the true progress of LM training.
Recent years witnessed significant progress in standardized benchmarking for English. Benchmarks such as GLUE, SuperGLUE, or KILT have become a~\emph{de facto} standard tools to compare large language models. Following the trend to replicate GLUE for other languages, the KLEJ benchmark\footnote{\emph{klej} is the word for glue in Polish} has been released for Polish. In this paper, we evaluate the progress in benchmarking for low-resourced languages. We note that only a handful of languages have such comprehensive benchmarks. We also note the gap in the number of tasks being evaluated by benchmarks for resource-rich English/Chinese and the rest of the world.

In this paper, we introduce \benchmarkname\footnote{\emph{lepiszcze} is the Polish word for glew, the Middle English predecessor of glue}, a new, comprehensive benchmark for Polish NLP with a large variety of tasks and high-quality operationalization of the benchmark.
We design \benchmarkname with flexibility in mind. Including new models, datasets, and tasks is as simple as possible while still offering data versioning and model tracking. In the first run of the benchmark, we test 13 experiments (task and dataset pairs) based on the five most recent LMs for Polish. We use five datasets from the Polish benchmark and add eight novel datasets. As the paper's main contribution, apart from \benchmarkname, we provide insights and experiences learned while creating the benchmark for Polish as the blueprint to design similar benchmarks for other low-resourced languages.

\end{abstract}

\section{Introduction}

Lack of reproducibility is an infuriating problem in machine learning practice. The inability to reproduce evaluation results and conduct reliable model comparisons is usually related to poor code quality, unclear and cryptic selection of hyper-parameter values, the random introduction of multiple factors affecting classification performance, and lack of a well-defined evaluation protocol~\citep{Pineau2021}. These problems can be circumvented by encouraging people to use standardized and peer-reviewed evaluation environments. The rapid development of diverse language technology has increased the need for reliable evaluation environments.

The reproducibility issues are intensifying even stronger as more novel language models emerge each year. We have seen a remarkable progress on many language understanding tasks, from language modeling \citep{Brown2020, Rae2021, Hoffmann2022}, Named Entity Recognition \citep{li-etal-2020-dice, ye-etal-2022-packed}, Q\&A \citep{Lan2020ALBERT, NEURIPS2019_dc6a7e65}, or various text classification tasks \citep{peters-etal-2018-deep, bingyu-arefyev-2022-document} in recent years. Moreover, in the last decade, data-centric models have become the major direction in solving most problems in the NLP area. Researchers and industry experts focus more on curated datasets and their maintenance processes. Hence, benchmarking models based on many datasets and their various and constantly changing versions is a great challenge. 

\begin{figure}[h]
  \centering
  \includegraphics[width=\textwidth]{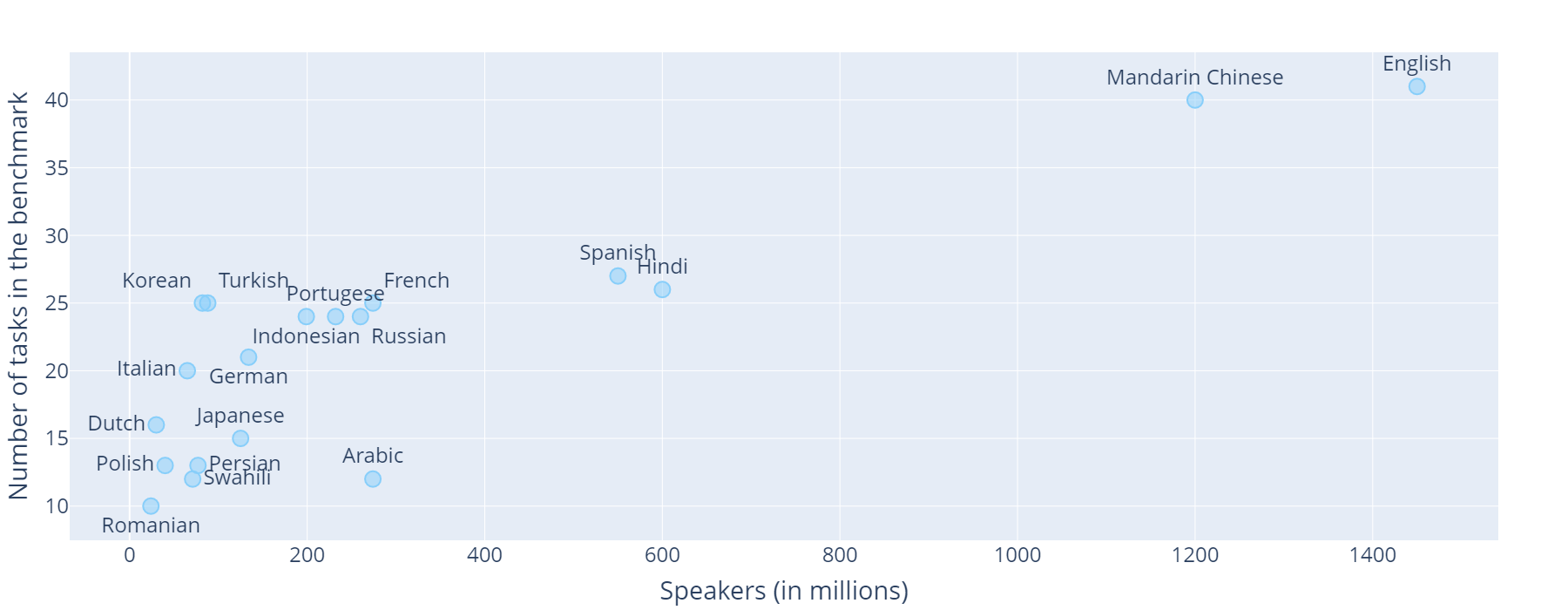}
  \caption{Tasks in large NLP benchmarks vs. the number of speakers, with Indian languages grouped for readability.}
  \label{fig:langs-vs-speakers}
\end{figure}

As shown in Figure~\ref{fig:langs-vs-speakers}, most NLP benchmarks are written for well-resourced languages such as English and Mandarin Chinese. This is understandable because many datasets exist in these languages, and many research teams are working in the context of these languages. Besides English and Mandarin Chinese, languages thoroughly covered with benchmarks include Indian, Spanish, French, and Portuguese. These are among the most commonly used languages in the world, and their position in the ranking is not surprising. However, we can also find Arabic and Japanese, widely spoken languages, but surprisingly few tasks are covered in benchmarks for these languages. Finally, we have languages with some benchmarks, such as Romanian, Persian, Dutch or Polish, but they only cover the most basic NLP tasks. 

In this work, we focus on Polish and provide datasets and tools to facilitate research on Polish NLP tasks. We designed the benchmarking process so that it could be easily applied to other languages. Thus, preparing and adding benchmarks for other low-resourced languages should become much less laborious.

The Polish benchmarking tradition has a relatively short history. One of the few platforms for evaluating and comparing modern language models for Polish is the KLEJ benchmark~\citep{klej}, a single-metric benchmark defined over a limited dataset. This simple practice for evaluating a model's performance no longer works. Current recommendations for the comparative evaluation of LMs advocate the inclusion of diversified tasks, challenges, and tests. Hence, we wanted to rethink and design a benchmark and environment to assess models so that they can still serve as valuable progress indicators.

Our main contributions are as follows:

\begin{itemize}
    \item We propose \benchmarkname, a new, extensive benchmark for Polish NLP with a large variety of tasks, expanding the previous Polish benchmark KLEJ with eight new datasets, published as a unified modern API.
    \item We design the benchmark and its maintenance using the best practices found in the literature, and we also investigate some of the most problematic aspects of creating benchmarks. We~share the lessons learned while building the benchmark.
    \item We present the summary of training and evaluation of more than 6000 different models for \benchmarkname, storing all information about code, dataset versions, parameters, metrics, predictions, or even information about their experimental environment.
\end{itemize}

\section{Related Work}
We used Google Scholar to review available NLP benchmarks for languages with at least 10 million speakers without going deeper into dialects (i.e., German includes all German dialects without dividing into Standard or Bavarian German). We searched for \texttt{<language name> NLP|NLG benchmark}. Furthermore, we dismissed benchmarks consisting of a single dataset. As a result, we found 35 benchmarks (see Table 1 in the Appendix), 
which included a total of 71 different tasks. Out of those, only 34 appeared in one language. These can be divided into two groups: specialized tasks which require a larger effort to build a good dataset (like diagnosis normalization, see \citep{WANG2020103418}), or misdefined tasks such as Named Entity Recognition in the Polish KLEJ benchmark, which was not a span labeling task, but rather a text fragment classification task to detect if it contains an entity, without providing the span. We provide more detailed results of our survey in the supplementary materials.

When it comes to language coverage, only 31 languages have an existing NLP or NLG benchmark out of 91 available in the 2022 edition of Ethnologue \citep{ethno2022}: Arabic, Assamese, Bengali, Chinese, Dutch, English, French, German, Gujarati, Hindi, Indonesian, Italian, Japanese, Kannada, Korean, Malayalam, Marathi, Odia, Persian, Polish, Portuguese, Punjabi, Romanian, Russian, Spanish, Swahili, Tamil, Telugu, Turkish, Urdu. The 74 tasks were not equally distributed per language, per Figure \ref{fig:langs-vs-speakers}.
Benchmarks for the two most commonly spoken languages: English and Chinese, would cover around 40 tasks. In contrast, the languages with the lowest number of tasks available in benchmarks and lower numbers of native speakers were Romanian and Polish (around 10 tasks). The results of our analysis are attached in Appendix.

The Polish language has a disproportionately small number of tasks in its main NLP benchmarks given nearly 40 million native speakers. KLEJ benchmark originally provided only 9 tasks, marking Polish the least task-covered European language concerning modern NLP and NLG benchmarked tasks. 
Once we take a deeper look at how tasks are formulated in KLEJ, we must acknowledge that the number of tasks formulated in a manner established in a given NLP sub-field is even smaller. There are two similar tasks of online reviews sentiment analysis, differing only in domains (PolEmo), and another sentiment analysis task (AR) framed as a regression task. Thus, the number of tasks can be reduced to 7 if we consider these datasets as one task.
Moreover, some of the tasks are ill-defined.
The NER task in KLEJ is not a sequence tagging but a document classification task. Summarization in KLEJ is evaluated based on classifying pairs of text and summary, the task is to predict whether the summary summarizes the text. Most benchmarks would define summarization as an NLG task, where the model is expected to generate the summary and would be evaluated with ROGUE or BLEU measures. A similar situation is happening in the Q\&A task. 

The KLEJ benchmark was created in 2020. Since then, no new datasets have been added, and the benchmark can be considered a little stale. Some of the tasks in KLEJ are not very difficult and diverse. Another potential problem with KLEJ is that it does not provide any environment for testing or submitting the model, as the submission requires only a prediction file. Finally, the heterogeneous, task-specific metrics for all the tasks in KLEJ could also be problematic when comparing the models as it may lead to erroneous conclusions. In our work, we aim to address these limitations. In particular, we plan to develop and maintain the long-term benchmark as part of the CLARIN-PL-Biz project. 

\section{\benchmarkname}

\benchmarkname 
is an open-source benchmark and a continuous-submission leaderboard, concentrating public Polish datasets (existing and new) in speciﬁc tasks. Integrating datasets and tasks with model performance and efﬁciency allows academia and industry to gauge performance on tasks of interest quickly. Finally, it intends to foster constructive competition and innovation by bringing together and promoting previously disparate resources. 

Our benchmark is structured into \textit{datasets}, \textit{tasks}, and \textit{models}. We design the benchmark to be easily extendable and flexible so that leaderboards for various subsets of datasets, tasks, and models can be added in the future. 

\subsection{Benchmark designing and construction process} \label{sec:lessons}
In this section, we introduce the benchmark construction process and lessons learned during this procedure, hoping that they could serve as a guide for other researchers that will face the task of creating a benchmark. The general design concept we follow is to make the submission process straightforward and benchmark easily extendable to new models and datasets, guaranteed by accessible experimental infrastructure and a unified submission procedure.
Our approach makes it effortless to test and compare models in a reliable and transparent way with the possibility of quickly entering new data. 

\paragraph{Task diversity}
The value of the benchmark depends directly on the chosen tasks and their diversity. If an unrepresentative collection of data and tasks is used to create a benchmark, the evaluation is of limited informative value for the further development of language models. If a benchmark consists only of closely related datasets, we can evaluate only a narrow part of the model's capabilities. Hence, one of the first and the most critical tasks for us was to gather many diverse tasks for Polish that cover different domains and tasks. We wanted to cover also many sources of text data in our benchmarking environment. The model’s performance could differ for books, social media, and other domain texts. Thus, having a representative collection of text data allows for evaluating the models in terms of their in-domain and out-of-domain generalization abilities. However, since Polish is a low-resource language, our choice was limited and we ended up with text classification, natural language inference, and sequence labeling tasks. We considered a few datasets for the regression task (e.g., CDSC-R or Allegro Reviews from KLEJ). However, in our opinion, mentioned datasets are unsuitable for benchmarking. It appears that annotation consistency is much lower than claimed in the original papers \citep{klej, wroblewska-krasnowska-kieras-2017-polish}. However, we hope to extend the benchmark with other than the mentioned above regression task in the future.

\paragraph{Dataset selection}
To select proper datasets for our benchmark, we first looked at the datasets available in KLEJ \citep{klej}. Many datasets have been described in their research papers, but they were still quite hard to obtain, and of course, they were in different formats. We also noticed that some tasks defined in the KLEJ benchmark were ill-defined, e.g., the NER task as text classification instead of sequence labeling. We set out to fix these problems and widen the scope of covered tasks and domains. Actual sequence tagging datasets (KPWr-NER and NKJP POS tagging) were added to the benchmark. 
Regarding the classification task, we added aspect-based sentiment analysis, political advertising detection, and punctuation restoration datasets to cover more diversified tasks. 
We also prepared two new datasets concerning legal text (Political Abusive Clauses) and dialogue systems (DiaBiz.Kom). 
It is important to mention that almost all of the datasets chosen by us (i.e., KPWr-NER, AspectEmo, PolEmo, DiaBiz.Kom, PAC, Political Advertising, and PSC) have been created by researchers in the \clarinpl  project; hence the annotation processes and inter-annotator agreements are properly described in the relevant papers.
The complete list of datasets with a short description can be found in Section \ref{sec:benchmark_datasets}. 

We challenged an interesting problem when extending the collection of datasets covered in the benchmark, namely, should we add a dataset that is not free and publicly available? It is an important choice when designing the benchmark. On the one hand, it contradicts the guidelines of open science, but on the other hand, it makes the benchmark more challenging and practically useful. After a lot of deliberations, we have decided to add to the benchmark the Dialogue Acts dataset --- DiaBiz.Kom (more in Section \ref{sec:dialogue-acts-dataset}) which is available only for internal usage of \clarinplbiz associates. Still, the dataset covers a significant collection of infrequent domain data for Polish targeted at spoken language understanding. The results of modern language models for this dataset present much room for improvement --- see Table \ref{tab:results}. 

\paragraph{Model selection}
The next step was to select initial models in the benchmark to provide a baseline and allow for easy comparison with the already published models. Each baseline model had to be available in the HuggingFace repository, and it could not be too big since we were again limited by the amount of available compute.\footnote{Maximum size of the model to perform hyperparameter search in a reasonable time was 350 million parameters} 
We first took the models of different sizes provided in KLEJ -- HerBERT-large is a top-1 model in the KLEJ benchmark. We used these models and our hyperparameter search module to validate the experimental setup and generate the first results for benchmark. We then utilized another popular mono-lingual encoder model and tested it against the previous one. Finally, to provide some diversity, we took the sentence model (which in fact is a cross-lingual) and ran the experiment. We plan to add new models to the benchmark to allow comparison of cross- vs. single-lingual and sentence- vs. word encoders. Table \ref{tab:lms} shows the final collection of models used for experiments.

\paragraph{Choice of metrics}
Every benchmark has to provide task-specific evaluation metrics. Even though we can focus on a single metric for a specific dataset and task in most cases, it could not be enough for many scenarios. A single metric can be insufficient to capture the varying cost of errors in many tasks. For instance, in sentiment analysis, the misclassification of "strong negative" as "negative" is far less consequential than the misclassification of the same case as "positive". 
In general, instead of focusing on a single metric, we want to evaluate language models using multiple metrics (and allow for the construction of compound weighted metrics). 
\benchmarkname calculates many different metrics, namely, F1-score, recall, precision, and accuracy. We allow sorting submissions based on various metrics. We believe that comparing models between tasks using different metrics may be misleading, which is why we use homogeneous metrics to compare and score the models. 

Many benchmarks concentrate only on metrics that evaluate the quality of predictions, without considering the cost of prediction. We believe that the omission of externalities, such as the computational requirements of models, leads to very biased rankings. The costs incurred by modern language models (i.e. in terms of their carbon footprint) can be significant and should be included in the evaluation. We track all Python environment information and completion times of all experiments for the benchmark. This way, we can build a custom leaderboard that incorporates computational costs as well. As \citep{ethayarajh-jurafsky-2020-utility} said, a highly inefficient model would provide less utility to practitioners but not to a leaderboard since it is a cost that only the former must bear.

\paragraph{Experimental environment}
We use several tools in our experimental environment to facilitate the use of the benchmark.
Benchmarking involves running many experiments and tracking their performance. We actively utilize the \textit{HuggingFace} repository to make the process of adding and testing datasets and models convenient. We unified all datasets into one accessible and easy-to-process data format, uploaded them to the HuggingFace Datasets~\footnote{\url{https://huggingface.co/datasets}} repository and ran all experiments using the HuggingFace hub. Technically, our benchmark allows us to choose any dataset or collect utterly new data, prepare data-loading scripts compatible with the HuggingFace platform, and evaluate models in the target language. 
We also develop our library \libraryname to unify the whole process of training, validating, performing hyperparameter search, testing, and submitting results to the leaderboard almost automatically, in only a few lines of code. 
We~believe that this step is crucial to allow continuous benchmarking and encourage researchers to submit their models or datasets.
Integrating benchmark libraries with HuggingFace Datasets platform opens new possibilities to evaluate language models in a multilingual zero-shot setting for any low-resource language. We~believe using a unified dataset inventory will contribute to the sustainable development of reliable evaluation data. 
The \libraryname  library is built on \textit{PyTorch, PyTorch Lightning} and \textit{Transformers} is easily extendable and modifiable; we plan to keep developing and expanding it over time. While the KLEJ benchmark required only a .zip file submission with predictions, we provide a~great level of technical support to the user with the extensive experimental environment that reinforces the reproducibility.

\paragraph{Standard splits problems}


Many benchmarks, such as GLUE and derived works, do not reveal test sets on which the benchmarking platform calculates the final results. Using static data splits leads to over-fitting and results in quick benchmark saturation. An alternative approach is based on multiple splits\citep{Gorman2020}, allowing for evaluation of the model's performance based on many different data partitions. In our evaluation pipelines, we follow this methodology and use not only a single train, dev, and test split but multiple splits. In our benchmark, we decided to implement a new experiment to evaluate non-original splits in the next benchmark version.

\paragraph{Continuous benchmarking}

The disadvantage of multi-task benchmarks such as GLUE, SuperGLUE, KLEJ, etc., is their lack of dynamics. The static benchmarks become quickly outdated and, therefore, useless from a practical perspective. As part of the \clarinplbiz\footnote{\url{http://clarin-pl.eu/index.php/en/home/}}, we plan to add more datasets, tasks, and models and maintain \benchmarkname benchmark continuously. We encourage other associated researchers to publish datasets and models in our benchmark. Many \benchmarkname datasets have been added with their source and author contributions. We track all model parameters and versions of each dataset. Hence, we can create a leaderboard for a specific version of the dataset in our benchmark.

\subsection{Datasets in the benchmark}
\label{sec:benchmark_datasets}
In this section, we briefly describe the final collection of datasets selected for the initial version of our benchmark. As of now, we also preserved the original splits of these datasets. 


\paragraph{PAC --- Polish Abusive Clauses Dataset} ''I have read and agree to the terms and conditions'' is one of the biggest lies on the Internet. Consumers rarely read the contracts they are required to accept. 
On the Internet, we probably skip most of the Terms and Conditions. However, we must remember that we have concluded many more contracts. 
European consumer law aims to prevent businesses from using so-called ``unfair contractual terms'' in their unilaterally drafted contracts, requiring consumers to accept. The PAC aims to detect ``unfair contractual term'' as the equivalent of an abusive clause. The task was formulated as binary text classification. The dataset has been created with the Office of Competition and Consumer Protection.
This dataset uses more than 700 contracts and gathers 4,129 examples of abusive clauses and 5,127 non-abusive contract fragments. 

It is worth noticing that the PAC dataset is important from an ethical point of view. Particularly it is based on actual agreements. The Office of Competition and Consumer Protection employees have checked the dataset to see if it contains Personal Identifiable Information (PII). A couple of such examples have been removed from the texts.

\paragraph{AspectEmo} Corpus \citep{Koco2021AspectEmoMC} is an extended version of the publicly available PolEmo 2.0 corpus. The AspectEmo corpus consists of 1,465 online customer reviews from the following domains: school, medicine, hotels, and products. All documents are annotated at the aspect level with six sentiment categories: strong negative, weak negative, neutral, weakly positive, and strong positive.

\paragraph{CDSC-E} Compositional Distributional Semantics Corpus \citep{wroblewska-krasnowska-kieras-2017-polish} is an entailment classification task. It consists of 1000 pairs of sentences and human-annotated entailment labels for each pair. There are three possible classes: entailment, contradiction, and neutral.

\paragraph{Dialogue Acts --- DiaBiz.Kom}
\label{sec:dialogue-acts-dataset}

It consists of 1,277,962 tokens in 1,104 transcribed call center phone conversations spanning eight domains. Each example is annotated by three linguists (in a 2+1 system, with an inter-annotator agreement of Positive Specific Agreement equal to 0.78 for annotation borders and categories and 0.86 for annotation borders) with dialogue acts in compliance with ISO 64217-2:2012 standard with layer of information concerning communicative functions. Within the benchmark, we consider the task of dialogue act classification, where each utterance is provided with its role in the dialogue. 
DiaBiz.Kom is annotation layer on top of DiaBiz \citep{pezik_lrec_2022} --- corpus  of Polish call center dialogues.

\paragraph{DYK} Did You Know (pol. Czy wiesz?) is a dataset that consists of 4,721 human-annotated question-answer pairs. It was simplified by \citep{klej} to binary classification to label denoting if the answer contained in the Wikipedia article is factually correct in light of the stated question.

\paragraph{KPWr-NER} is a part of the Polish Corpus of Wrocław University of Technology (Korpus Języka Polskiego Politechniki Wrocławskiej)  \citep{broda-etal-2012-kpwr} is a named entity recognition dataset focusing on fine-grained categories of entities (82 classes) using BIO notation. It contains 13,959 training and 4,323 test human annotated sentences, originating from texts covering a large variety of domains, genres, and sources.

\paragraph{NKJP-POS} is a part of the National Corpus of Polish (Narodowy Korpus Języka Polskiego)
 \citep{przepiorkowski_narodowy_2012}. Its objective is the part-of-speech tagging task. The dataset  contains 85,663 sentences tagged with 35 tags. During the creation of the corpus, texts were annotated by humans from various sources, covering many domains and genres.

\paragraph{PolEmo 2.0}  \citep{kocon-etal-2019-multi} is a dataset of online consumer reviews from four domains: medicine, hotels, products, and university. It consists of 8,216 reviews having 57,466 sentences. The aim is to predict one of the sentiment classes: positive, negative, neutral, or ambiguous. During the development of the KLEJ benchmark \citep{klej} two tasks that differ in the context used during evaluation have been created: \textbf{in-domain} and \textbf{out-domain}. In contrast, we preserved the original data split and utilized all domains.

\paragraph{Political Advertising} dataset \citep{augustyniak-etal-2020-political} aims for detecting specific text chunks and categories of political advertising in the Polish language. It contains 1,705 human-annotated tweets tagged with nine categories, constituting campaigning under Polish electoral law. The authors achieved 0.65 inter-annotator agreement (Cohen's kappa score) for the sequence labeling task, and they used an additional annotator to resolve the mismatches between the first two annotators, improving the final consistency of annotations.

\paragraph{PSC}
Polish Summaries Corpus  \citep{ogrodniczuk-kopec-2014-polish} consists of 569 news summaries done by human annotators. We used the simplified version developed by \citep{klej} for the purpose of the KLEJ benchmark. They formulated a binary paraphrase classification task by matching positive and negative pairs using the procedure detailed in the publication. 

\paragraph{Punctuation Restoration} is a crowd-sourced text and audio dataset of Polish Wikipedia pages read out loud by Polish lectors. The base dataset is divided into  conversational (WikiTalks) and information (WikiNews) parts. Then the texts were read by hundred people, which resulted in 36 hours of transcription. Punctuation restoration includes 1000 texts - 800 trains and 200 test examples. This dataset is part of PolEval 2021 Competition \footnote{\url{http://2021.poleval.pl/tasks/}}.

\begin{table}[ht]
    \small
    \caption{Datasets available in the \benchmarkname benchmark with sizes of the train, dev, and test sets. The datasets that were previously incorporated into the KLEJ benchmark are marked with \textcolor{red}{*} symbol. \textbf{WIP} denotes the dataset for which we present preliminary results.}
    \centering
    \begin{tabular}{lllrrrr}
    \toprule
    Name & Domain & Task & Train & Dev & Test & \#Classes\\
    \midrule
    \makecell[l]{CDSC-E\textcolor{red}{*}}                    & image captions                & \makecell[l]{Entailment\\ Classification}             & 8000 & 1000 & 1000 & 3\\
    \makecell[l]{DYK\textcolor{red}{*}}                      & Wikipedia                     & Q\&A Classification                   & 4154 & 0 & 1029 &  2\\
    \makecell[l]{PolEmo 2.0\\(In-Domain)\textcolor{red}{*}}    & online reviews                & Sentiment Analysis                    & 5783 & 723 & 722 &  4\\
    \makecell[l]{PolEmo 2.0\\(Out-Domain)\textcolor{red}{*}}  & online reviews                & Sentiment Analysis                    & 5783 & 494 & 494 &  4\\
    \makecell[l]{PSC\textcolor{red}{*}}                  & news                          &  \makecell[l]{Paraphrase\\Classification}               & 4302 & 0 & 1078  & 4\\
    \midrule
    \makecell[l]{Abusive\\Clauses }                            & legal texts                   & Abusive Clauses Detection                                      &  4284  & 1519 & 3453 & 2 \\
    \makecell[l]{AspectEmo}                                   & online reviews                & \makecell[l]{Aspect-based\\Sentiment Analysis}       & 1173 & 0 & 292  & 7\\
    \makecell[l]{KPWr NER}                                    & misc.                         & NER              & 13959 & 0 & 4323 &  82\\
    \makecell[l]{NKJP POS}                                    & misc.                         & POS Tagging                & 78219 & 0 & 7444 &  35\\
    \makecell[l]{PolEmo 2.0}                                  & online reviews                & Sentiment Analysis                    & 6573 & 823 & 820  &  4\\
    \makecell[l]{Political\\Advertising}                       & social media                  & \makecell[l]{Political Advertising \\Detection}                                      &  1020  & 340 & 341 & 9 \\
    \makecell[l]{Punctuation\\Restoration}                     & \makecell[l]{Wikipedia Talk,\\Wikinews}      & \makecell[l]{Punctuation\\Restoration}               & 800 & 0 & 200  & 8 \\
    \makecell[l]{Dialogue Acts\\(WIP)}                                  & \makecell[l]{call center phone\\ conversations}  & \makecell[l]{Dialogue Acts\\ Classification} & 70454 & 8807 & 8807  &  54\\
    \bottomrule
    \end{tabular}
        \label{tab:datasets}
\end{table}

\section{Experiments}
\label{sec:experiments}

We conducted experiments on 13 datasets and five language models. Each language model was fine-tuned for a given dataset and was evaluated separately. Experiments were performed with our developed library \libraryname, which provides predefined pipelines for text classification, text pairs classification, and sequence labeling. To ensure reproducibility of our experiments, we utilized MLOps tools: Data Version Control (DVC) \citep{ruslan_kuprieiev_2022_7020480} for pipelines and tracking of datasets and models; Weight\&Biases \citep{wandb} for experiments summaries and metrics tracking. We share the code of our experiments on GitHub repository \footnote{\url{\repositoryurl}} and model tracking dashboard \footnote{\url{\weightbiasesurl}}.

\subsection{Initial models for benchmark}
We picked four recent transformer-based language models for Polish publicly available in the HuggingFace hub, along with one multilingual XLM-RoBERTa model. We present those models with their total number of parameters and repository location in Table \ref{tab:lms}. For fine-tuning, we utilize sequence and token classification models from the transformers library~\citep{wolf-etal-2020-transformers}, consisting of a single linear classification layer with dropout. For the initial datasets evaluation, we chose both cased and uncased versions of the language models. 

\begin{table}[ht]
    \small
    \caption{Language Models used for experiments. All models can be accessed via the HuggingFace repository.}
    \centering
    \begin{tabular}{c|c|c}
        \toprule
         Model &  \#Params &   HuggingFace Repository Name\\
         \midrule
         PolBERT (base, cased), \citep{Kleczek2020} & 132M & \texttt{dkleczek/bert-base-polish-cased-v1} \\
         PolBERT (base, uncased), \citep{Kleczek2020} & 132M & \texttt{dkleczek/bert-base-polish-uncased-v1} \\
         \makecell{HerBERT (base, cased)\\ \citep{mroczkowski-etal-2021-herbert}} & 124M &  \texttt{allegro/herbert-base-cased} \\
         \makecell{HerBERT (large, cased)\\ \citep{mroczkowski-etal-2021-herbert}} & 355M & \texttt{allegro/herbert-large-cased} \\
         \makecell{XLM-RoBERTa (paraphrase)\\
         \citep{reimers-2019-sentence-bert}} & 278M & \makecell{\texttt{sentence-transformers/paraphrase-}\\\texttt{xlm-r-multilingual-v1}} \\
         \bottomrule
    \end{tabular}
    \label{tab:lms}
\end{table}

\subsection{Hyper-parameter search (HPS)}
To fairly compare different transformer models across various tasks, we performed a hyper-parameter search (HPS) to obtain the best configuration for fine-tuning the language model to a particular task. We performed a hyper-parameter search separately for each combination of tasks and language models (which we restricted to 100 iterations). Under the hood, we utilized the Optuna framework \citep{akiba2019optuna} wrapper from the \libraryname library. We also logged each run in the hyper-parameter search via Weights\&Biases PyTorch Lightning logger.

Experiments were computed using a server with five Titan RTX GPU cards. We logged over 6000 runs in the Weight\&Biases dashboard, which took over 2000 hours to complete. We reported metrics, hyper-parameters, dataset information, and package versions in each run.

We used macro averaged F1 measure as the metric for the objective function. Evaluation of models in the hyper-parameter search stage was performed on the validation subset of the dataset. In case validation subset was missing, we randomly sampled 10\% of the training subset. After obtaining the best hyper-parameter configuration, we no longer need such a subset, so we use original subsets for the final model evaluation.

For each dataset and language model pair, we choose the best configuration from the HPS process in terms of the best F1-macro score on the validation subset. We retrain models five times and calculate various metrics on test sets such as accuracy, precision, recall, and F1 with different averaging (micro, macro, weighted) and class or tags metrics (accuracy, precision, recall, and F1).

\subsection{Results of an initial set of trained models}

Evaluation results are presented in Table \ref{tab:results}, where we report macro averaged F1 metric for each dataset. Other metrics results can be found in the appendix Section A.3.
As we observe the performance of models above 80\% in text classification datasets (except out-domain dataset), these models perform poorly considering most sequence tagging tasks. Even the best performing model (HerBERT, large) shows F1-macro around 39\% for AspectEmo and 46\% in Punctuation Restoration. Considering those results, we can state that we still need better models that can cope with complex and under-represented tasks. Multilingual models XLM-RoBERTa and HerBERT (large) are comparable only for one dataset (around 2 percentage points difference for the Abusive Clauses dataset). However, for other tasks, the gap is much bigger, and it fails even more in sequence tagging tasks with a difference of up to 70\%. 

We also reported preliminary results for Dialogue Acts classification tasks. For the tested language model, we achieved a comparable performance of about 50\%. Considering limited computational resources and fine-tuning schemes, these results may be updated in our future work.

\begin{table}[ht!]
\centering
\small
\caption{
 Macro F1 performance of evaluated models on the test subsets. We present values as the mean and standard deviations over 5 model retrains. The mean rank row is the average of a~ranking established on the mean of model retrains. Values marked with \textbf{Bold} present the best results for a~single dataset. Additionally, we indicate datasets previously appeared in the KLEJ benchmark with \textcolor{red}{*}. \textbf{WIP} denotes the dataset for which we present preliminary results.
}
\label{tab:results}
\begin{tabular}{lrrrrr}
\toprule
 & \makecell{HerBERT\\(base, cased)} & \makecell{HerBERT\\(large, cased)} & \makecell{PolBERT\\(base, cased)} & \makecell{PolBERT\\(base, \\uncased)} & \makecell{XLM-\\RoBERTa\\(paraphrase)} \\
 
\midrule
\makecell[l]{CDSC-E\textcolor{red}{*}} & $\mathBF{90.96 \pm 0.73}$ & $90.48 \pm 0.20$ & $88.95 \pm 0.31$ & $90.62 \pm 0.27$ & $82.62 \pm 0.88$ \\
\makecell[l]{DYK\textcolor{red}{*}} & $\mathBF{82.39 \pm 1.43}$ & $79.58 \pm 0.59$ & $75.87 \pm 0.98$ & $74.41 \pm 1.15$ & $58.93 \pm 7.98$ \\
\makecell[l]{PolEmo 2.0 \\In-Domain\textcolor{red}{*}} & $88.10 \pm 0.36$ & $\mathBF{88.34 \pm 0.63}$ & $85.32 \pm 0.45$ & $85.71 \pm 0.40$ & $83.75 \pm 0.45$ \\
\makecell[l]{PolEmo 2.0 \\Out-Domain\textcolor{red}{*}} & $\mathBF{57.31 \pm 2.93}$ & $57.08 \pm 2.03$ & $54.10 \pm 3.82$ & $54.29 \pm 1.83$ & $45.12 \pm 3.40$ \\
\makecell[l]{PSC\textcolor{red}{*}} & $97.90 \pm 0.24$ & $98.33 \pm 0.69$ & $\mathBF{98.95 \pm 0.13}$ & $98.87 \pm 0.10$ & $58.85 \pm 1.49$ \\
\midrule
\makecell[l]{Abusive \\Clauses} & $85.66 \pm 0.58$ & $\mathBF{86.57 \pm 0.91}$ & $85.93 \pm 0.66$ & $85.74 \pm 0.86$ & $84.32 \pm 0.71$ \\
\makecell[l]{AspectEmo}  & $37.28 \pm 0.71$ & $\mathBF{39.44 \pm 1.74}$ & $30.01 \pm 0.58$ & $31.48 \pm 1.06$ & $18.42 \pm 0.98$ \\
\makecell[l]{KPWr NER} & $\mathBF{54.22 \pm 0.76}$ & $52.68 \pm 1.39$ & $48.01 \pm 0.76$ & $40.21 \pm 0.50$ & $36.13 \pm 0.44$ \\
\makecell[l]{NKJP POS} & $94.59 \pm 0.56$ & $\mathBF{96.14 \pm 0.38}$ & $94.34 \pm 0.61$ & $94.54 \pm 0.19$ & $90.29 \pm 0.51$ \\
\makecell[l]{PolEmo 2.0} & $86.78 \pm 0.79$ & $\mathBF{89.33 \pm 0.49}$ & $85.89 \pm 1.25$ & $85.83 \pm 0.47$ & $84.12 \pm 0.47$ \\
\makecell[l]{Political \\Advertising}  & $61.42 \pm 1.38$ & $62.16 \pm 0.14$ & $58.94 \pm 1.92$ & $\mathBF{62.52 \pm 1.23}$ & $56.68 \pm 0.94$ \\
\makecell[l]{Punctuation \\Restoration} & $45.59 \pm 0.38$ & $\mathBF{46.68 \pm 0.61}$ & $38.89 \pm 0.91$ & $41.31 \pm 0.59$ & $14.33 \pm 1.94$ \\
\makecell[l]{Dialogue Acts\\(WIP)} & $49.54 \pm 0.74$ & $\mathBF{51.11 \pm 0.85}$ & $50.20 \pm 1.32$ & $48.87 \pm 0.90$ & $49.05 \pm 0.39$ \\
\midrule
Mean rank & 2.15 & 1.62 & 3.23 & 3.08 & 4.92 \\
\bottomrule 
\end{tabular}

\end{table}

\section{Limitations}
\label{sec:limitations}
This study has potential limitations which are listed below. 
First, we do not give a human baseline score to datasets in the benchmark as in SuperGLUE \citep{Wang2019}. Second, in the initial version of the benchmark, we do not solve the standard split problem and evaluate the model on predefined original splits, we work on the second batch of results with more focus on diversified splits. Finally, due to practical constraints, the initial version of the benchmark does not include baselines with static embeddings, but they will be added as well in the second batch of results. Our intention is to keep the benchmark dynamic; we plan to add new datasets with various tasks which were not covered in this version of the benchmark, including NLG tasks such as summarization or translation. 


\section{Conclusions and Future Work}
\label{sec:conclusions}

In this paper, we have introduced \benchmarkname, a new comprehensive benchmark for Polish NLP. \benchmarkname is characterized by the large variety of NLP tasks and high-quality operationalization of the benchmark. The benchmarking approach is designed to maximize the flexibility and portability of other low-resourced languages. Adding new models, datasets, or NLP tasks is simple and intuitive. The benchmark internally supports data versioning and model tracking for improved reproducibility. In the first run of the benchmark, we tested 13 experiments (task and dataset pairs) based on the five most recent LMs for Polish to prove the usability and usefulness of \benchmarkname. 

An important added value of the paper is sharing our experiences collected during the work on the benchmark. We hope that NLP researchers working on other low-resourced languages will find our comments and suggestions useful. Below we summarize the most important issues encountered during our work on \benchmarkname:

\begin{itemize}
    \item \emph{multiple metrics}: it is important to provide implementations of multiple metrics that can be measured and compared across NLP tasks and datasets; evaluating language models on a~single metric produces a~distorted view of models' capabilities,
    \item \emph{diversity trumps openness}: one should not refrain from including closed datasets in the evaluation; they prevent over-fitting (as LMs are unlikely to see these datasets during training) and provide a good estimation of LM's performance on difficult tasks,
    \item \emph{include prediction costs}: the quality of prediction as measured by traditional NLP metrics is not enough, for the benchmark to be practically useful, one must be able to compare the computational resources consumed by LMs as well,
    \item \emph{interface matters}: making the benchmark interface simple and conventional hugely improves its usefulness and the probability of wide adoption; in the space of NLP models, the \texttt{HuggingFace} is the obvious choice of interface blueprints.
\end{itemize}

We plan to add more Natural Language Understanding and Spoken Language Understanding tasks to the benchmark. We want to use our \libraryname library to evaluate language models, other contextual embeddings, and static word representations with simpler than transformer-based models. The critical direction in the benchmark is to run experiments not only with the predefined data splits~\citep{Gorman2020} but also to use other splits to check the model's robustness properly. We hope that this benchmark will encourage the scientific community to work in transparent and reproducible environments, leading to a rapid improvement of the current language technology for the Polish language.

\begin{ack}

The work was partially supported by (1) the Polish Ministry of Education and Science, CLARIN-PL; (2) the European Regional Development Fund as a part of the 2014-2020 Smart Growth Operational Programme, CLARIN -- Common Language Resources and Technology Infrastructure, (3) project CLARIN-Q (agreement no. 2022/WK/09), and (4) the Department of Artificial Intelligence at Wroclaw University of Science and Technology.

\end{ack}

\bibliographystyle{acl_natbib}
\bibliography{references}

\begin{thebibliography}{64}
\expandafter\ifx\csname natexlab\endcsname\relax\def\natexlab#1{#1}\fi

\bibitem[{Akiba et~al.(2019)Akiba, Sano, Yanase, Ohta, and
  Koyama}]{akiba2019optuna}
Takuya Akiba, Shotaro Sano, Toshihiko Yanase, Takeru Ohta, and Masanori Koyama.
  2019.
\newblock Optuna: A next-generation hyperparameter optimization framework.
\newblock In \emph{Proceedings of the 25th ACM SIGKDD international conference
  on knowledge discovery \& data mining}, pages 2623--2631.

\bibitem[{Augustyniak et~al.(2020)Augustyniak, Rajda, Kajdanowicz, and
  Bernaczyk}]{augustyniak-etal-2020-political}
Lukasz Augustyniak, Krzysztof Rajda, Tomasz Kajdanowicz, and Micha{\l}
  Bernaczyk. 2020.
\newblock \href {https://www.aclweb.org/anthology/2020.winlp-1.28} {Political
  advertising dataset: the use case of the polish 2020 presidential elections}.
\newblock In \emph{Proceedings of the The Fourth Widening Natural Language
  Processing Workshop}, pages 110--114, Seattle, USA. Association for
  Computational Linguistics.

\bibitem[{Biewald(2020)}]{wandb}
Lukas Biewald. 2020.
\newblock \href {https://www.wandb.com/} {Experiment tracking with weights and
  biases}.
\newblock Software available from wandb.com.

\bibitem[{Bingyu and Arefyev(2022)}]{bingyu-arefyev-2022-document}
Zhang Bingyu and Nikolay Arefyev. 2022.
\newblock \href {https://doi.org/10.18653/v1/2022.insights-1.17} {The document
  vectors using cosine similarity revisited}.
\newblock In \emph{Proceedings of the Third Workshop on Insights from Negative
  Results in NLP}, pages 129--133, Dublin, Ireland. Association for
  Computational Linguistics.

\bibitem[{Blinov et~al.(2022)Blinov, Reshetnikova, Nesterov, Zubkova, and
  Kokh}]{blinov2022rumedbench}
Pavel Blinov, Arina Reshetnikova, Aleksandr Nesterov, Galina Zubkova, and
  Vladimir Kokh. 2022.
\newblock Rumedbench: A russian medical language understanding benchmark.
\newblock \emph{arXiv preprint arXiv:2201.06499}.

\bibitem[{Broda et~al.(2012)Broda, Marci{\'n}czuk, Maziarz, Radziszewski, and
  Wardy{\'n}ski}]{broda-etal-2012-kpwr}
Bartosz Broda, Micha{\l} Marci{\'n}czuk, Marek Maziarz, Adam Radziszewski, and
  Adam Wardy{\'n}ski. 2012.
\newblock \href
  {http://www.lrec-conf.org/proceedings/lrec2012/pdf/965_Paper.pdf} {{KPW}r:
  Towards a free corpus of {P}olish}.
\newblock In \emph{Proceedings of the Eighth International Conference on
  Language Resources and Evaluation ({LREC}'12)}, pages 3218--3222, Istanbul,
  Turkey. European Language Resources Association (ELRA).

\bibitem[{Brown et~al.(2020)Brown, Mann, Ryder, Subbiah, Kaplan, Dhariwal,
  Neelakantan, Shyam, Sastry, Askell, Agarwal, Herbert-Voss, Krueger, Henighan,
  Child, Ramesh, Ziegler, Wu, Winter, Hesse, Chen, Sigler, Litwin, Gray, Chess,
  Clark, Berner, McCandlish, Radford, Sutskever, and Amodei}]{Brown2020}
Tom~B. Brown, Benjamin Mann, Nick Ryder, Melanie Subbiah, Jared Kaplan,
  Prafulla Dhariwal, Arvind Neelakantan, Pranav Shyam, Girish Sastry, Amanda
  Askell, Sandhini Agarwal, Ariel Herbert-Voss, Gretchen Krueger, Tom Henighan,
  Rewon Child, Aditya Ramesh, Daniel~M. Ziegler, Jeffrey Wu, Clemens Winter,
  Christopher Hesse, Mark Chen, Eric Sigler, Mateusz Litwin, Scott Gray,
  Benjamin Chess, Jack Clark, Christopher Berner, Sam McCandlish, Alec Radford,
  Ilya Sutskever, and Dario Amodei. 2020.
\newblock \href {http://arxiv.org/abs/2005.14165} {{Language models are
  few-shot learners}}.
\newblock \emph{Advances in Neural Information Processing Systems},
  2020-Decem(NeurIPS).

\bibitem[{Cahyawijaya et~al.(2021)Cahyawijaya, Winata, Wilie, Vincentio, Li,
  Kuncoro, Ruder, Lim, Bahar, Khodra et~al.}]{cahyawijaya2021indonlg}
Samuel Cahyawijaya, Genta~Indra Winata, Bryan Wilie, Karissa Vincentio,
  Xiaohong Li, Adhiguna Kuncoro, Sebastian Ruder, Zhi~Yuan Lim, Syafri Bahar,
  Masayu Khodra, et~al. 2021.
\newblock Indonlg: Benchmark and resources for evaluating indonesian natural
  language generation.
\newblock In \emph{Proceedings of the 2021 Conference on Empirical Methods in
  Natural Language Processing}, pages 8875--8898.

\bibitem[{Canete et~al.(2022)Canete, Chaperon, Fuentes, Ho, Kang, and
  P{\'e}rez}]{canete2020spanish}
Jos{\'e} Canete, Gabriel Chaperon, Rodrigo Fuentes, Jou-Hui Ho, Hojin Kang, and
  Jorge P{\'e}rez. 2022.
\newblock Spanish pre-trained bert model and evaluation data.
\newblock In \emph{Proceedings of the Practical Machine Learning for Developing
  Countries @ ICLR 2022}.

\bibitem[{Chen et~al.(2022)Chen, Xu, Fu, Shi, Li, Zhang, Sun, Li, Xiao, and
  Zhou}]{chen2022kar}
Jiangjie Chen, Rui Xu, Ziquan Fu, Wei Shi, Zhongqiao Li, Xinbo Zhang, Changzhi
  Sun, Lei Li, Yanghua Xiao, and Hao Zhou. 2022.
\newblock E-kar: A benchmark for rationalizing natural language analogical
  reasoning.
\newblock In \emph{Findings of the Association for Computational Linguistics:
  ACL 2022}, pages 3941--3955.

\bibitem[{Conneau and Kiela(2018)}]{conneau2018senteval}
Alexis Conneau and Douwe Kiela. 2018.
\newblock Senteval: An evaluation toolkit for universal sentence
  representations.
\newblock \emph{arXiv preprint arXiv:1803.05449}.

\bibitem[{Dumitrescu et~al.(2021)Dumitrescu, Rebeja, Lorincz, Gaman, Avram,
  Ilie, Pruteanu, Stan, Rosia, Iacobescu et~al.}]{dumitrescu2021liro}
Stefan~Daniel Dumitrescu, Petru Rebeja, Beata Lorincz, Mihaela Gaman, Andrei
  Avram, Mihai Ilie, Andrei Pruteanu, Adriana Stan, Lorena Rosia, Cristina
  Iacobescu, et~al. 2021.
\newblock Liro: Benchmark and leaderboard for romanian language tasks.
\newblock In \emph{Thirty-fifth Conference on Neural Information Processing
  Systems Datasets and Benchmarks Track (Round 1)}.

\bibitem[{Ethayarajh and Jurafsky(2020)}]{ethayarajh-jurafsky-2020-utility}
Kawin Ethayarajh and Dan Jurafsky. 2020.
\newblock \href {https://doi.org/10.18653/v1/2020.emnlp-main.393} {Utility is
  in the eye of the user: A critique of {NLP} leaderboards}.
\newblock In \emph{Proceedings of the 2020 Conference on Empirical Methods in
  Natural Language Processing (EMNLP)}, pages 4846--4853, Online. Association
  for Computational Linguistics.

\bibitem[{Fallahnejad and Zarezade(2021)}]{persian-nlp-benchmark}
Zohreh Fallahnejad and Ali Zarezade. 2021.
\newblock Persian nlp benchmark.
\newblock
  \url{https://github.com/Mofid-AI/persian-nlp-benchmark#persian-nlp-benchmark}.

\bibitem[{Fennig et~al.(2022)Fennig, Eberhard, and Simons}]{ethno2022}
Charles Fennig, David Eberhard, and Gary~F. Simons, editors. 2022.
\newblock \href {http://www.ethnologue.com} {\emph{Ethnologue: Languages of the
  World}}, twenty-fifth edition.
\newblock SIL International, Dallas, TX, USA.

\bibitem[{Gehrmann et~al.(2021)Gehrmann, Adewumi, Aggarwal, Ammanamanchi,
  Aremu, Bosselut, Chandu, Clinciu, Das, Dhole et~al.}]{gehrmann2021gem}
Sebastian Gehrmann, Tosin Adewumi, Karmanya Aggarwal, Pawan~Sasanka
  Ammanamanchi, Anuoluwapo Aremu, Antoine Bosselut, Khyathi~Raghavi Chandu,
  Miruna-Adriana Clinciu, Dipanjan Das, Kaustubh Dhole, et~al. 2021.
\newblock The gem benchmark: Natural language generation, its evaluation and
  metrics.
\newblock In \emph{Proceedings of the 1st Workshop on Natural Language
  Generation, Evaluation, and Metrics (GEM 2021)}, pages 96--120.

\bibitem[{Gomes(2020)}]{Gomes2020}
J.~R.~S. Gomes. 2020.
\newblock Plue: Portuguese language understanding evaluation.
\newblock \url{https://github.com/jubs12/PLUE}.

\bibitem[{Gorman and Bedrick(2020)}]{Gorman2020}
Kyle Gorman and Steven Bedrick. 2020.
\newblock \href {https://doi.org/10.18653/v1/p19-1267} {{We need to talk about
  standard splits}}.
\newblock \emph{ACL 2019 - 57th Annual Meeting of the Association for
  Computational Linguistics, Proceedings of the Conference}, pages 2786--2791.

\bibitem[{Guan et~al.(2022)Guan, Feng, Chen, He, Mao, Fan, and
  Huang}]{guan2022lot}
Jian Guan, Zhuoer Feng, Yamei Chen, Ruilin He, Xiaoxi Mao, Changjie Fan, and
  Minlie Huang. 2022.
\newblock Lot: A story-centric benchmark for evaluating chinese long text
  understanding and generation.
\newblock \emph{Transactions of the Association for Computational Linguistics},
  10:434--451.

\bibitem[{Hoffmann et~al.(2022)Hoffmann, Borgeaud, Mensch, Buchatskaya, Cai,
  Rutherford, Casas, Hendricks, Welbl, Clark, Hennigan, Noland, Millican,
  Driessche, Damoc, Guy, Osindero, Simonyan, Elsen, Rae, Vinyals, and
  Sifre}]{Hoffmann2022}
Jordan Hoffmann, Sebastian Borgeaud, Arthur Mensch, Elena Buchatskaya, Trevor
  Cai, Eliza Rutherford, Diego de~Las Casas, Lisa~Anne Hendricks, Johannes
  Welbl, Aidan Clark, Tom Hennigan, Eric Noland, Katie Millican, George van~den
  Driessche, Bogdan Damoc, Aurelia Guy, Simon Osindero, Karen Simonyan, Erich
  Elsen, Jack~W. Rae, Oriol Vinyals, and Laurent Sifre. 2022.
\newblock \href {https://doi.org/10.48550/ARXIV.2203.15556} {Training
  compute-optimal large language models}.

\bibitem[{Hu et~al.(2020)Hu, Ruder, Siddhant, Neubig, Firat, and
  Johnson}]{hu2020xtreme}
Junjie Hu, Sebastian Ruder, Aditya Siddhant, Graham Neubig, Orhan Firat, and
  Melvin Johnson. 2020.
\newblock Xtreme: A massively multilingual multi-task benchmark for evaluating
  cross-lingual generalisation.
\newblock In \emph{International Conference on Machine Learning}, pages
  4411--4421. PMLR.

\bibitem[{Kakwani et~al.(2020)Kakwani, Kunchukuttan, Golla, Gokul,
  Bhattacharyya, Khapra, and Kumar}]{kakwani2020indicnlpsuite}
Divyanshu Kakwani, Anoop Kunchukuttan, Satish Golla, NC~Gokul, Avik
  Bhattacharyya, Mitesh~M Khapra, and Pratyush Kumar. 2020.
\newblock Indicnlpsuite: Monolingual corpora, evaluation benchmarks and
  pre-trained multilingual language models for indian languages.
\newblock In \emph{Findings of the Association for Computational Linguistics:
  EMNLP 2020}, pages 4948--4961.

\bibitem[{Khashabi et~al.(2021)Khashabi, Cohan, Shakeri, Hosseini, Pezeshkpour,
  Alikhani, Aminnaseri, Bitaab, Brahman, Ghazarian
  et~al.}]{khashabi2021parsinlu}
Daniel Khashabi, Arman Cohan, Siamak Shakeri, Pedram Hosseini, Pouya
  Pezeshkpour, Malihe Alikhani, Moin Aminnaseri, Marzieh Bitaab, Faeze Brahman,
  Sarik Ghazarian, et~al. 2021.
\newblock Parsinlu: A suite of language understanding challenges for persian.
\newblock \emph{Transactions of the Association for Computational Linguistics},
  9:1147--1162.

\bibitem[{Kiela et~al.(2021)Kiela, Bartolo, Nie, Kaushik, Geiger, Wu, Vidgen,
  Prasad, Singh, Ringshia et~al.}]{kiela2021dynabench}
Douwe Kiela, Max Bartolo, Yixin Nie, Divyansh Kaushik, Atticus Geiger,
  Zhengxuan Wu, Bertie Vidgen, Grusha Prasad, Amanpreet Singh, Pratik Ringshia,
  et~al. 2021.
\newblock Dynabench: Rethinking benchmarking in nlp.
\newblock In \emph{Proceedings of the 2021 Conference of the North American
  Chapter of the Association for Computational Linguistics: Human Language
  Technologies}, pages 4110--4124.

\bibitem[{Kim et~al.(2022)Kim, Jang, Kwon, and Davis}]{kim2022kobest}
Dohyeong Kim, Myeongjun Jang, Deuk~Sin Kwon, and Eric Davis. 2022.
\newblock Kobest: Korean balanced evaluation of significant tasks.
\newblock \emph{arXiv preprint arXiv:2204.04541}.

\bibitem[{Koco{\'n} et~al.(2019)Koco{\'n}, Mi{\l}kowski, and
  Za{\'s}ko-Zieli{\'n}ska}]{kocon-etal-2019-multi}
Jan Koco{\'n}, Piotr Mi{\l}kowski, and Monika Za{\'s}ko-Zieli{\'n}ska. 2019.
\newblock \href {https://doi.org/10.18653/v1/K19-1092} {Multi-level sentiment
  analysis of {P}ol{E}mo 2.0: Extended corpus of multi-domain consumer
  reviews}.
\newblock In \emph{Proceedings of the 23rd Conference on Computational Natural
  Language Learning (CoNLL)}, pages 980--991, Hong Kong, China. Association for
  Computational Linguistics.

\bibitem[{Kocoń et~al.(2021)Kocoń, Radom, Kaczmarz-Wawryk, Wabnic,
  Zajączkowska, and Zaśko-Zielińska}]{Koco2021AspectEmoMC}
Jan Kocoń, Jarema Radom, Ewa Kaczmarz-Wawryk, Kamil Wabnic, Ada Zajączkowska,
  and Monika Zaśko-Zielińska. 2021.
\newblock Aspectemo: Multi-domain corpus of consumer reviews for aspect-based
  sentiment analysis.
\newblock \emph{2021 International Conference on Data Mining Workshops
  (ICDMW)}, pages 166--173.

\bibitem[{Koto et~al.(2020)Koto, Rahimi, Lau, and Baldwin}]{koto2020indolem}
Fajri Koto, Afshin Rahimi, Jey~Han Lau, and Timothy Baldwin. 2020.
\newblock Indolem and indobert: A benchmark dataset and pre-trained language
  model for indonesian nlp.
\newblock In \emph{Proceedings of the 28th International Conference on
  Computational Linguistics}, pages 757--770.

\bibitem[{Kuprieiev et~al.(2022)Kuprieiev, skshetry, Petrov, Redzyński,
  Rowlands, da~Costa-Luis, Schepanovski, Shcheklein, Taskaya, Gao, Orpinel,
  de~la Iglesia~Castro, Santos, Sharma, Zhanibek, Hodovic, Berenbaum, Kodenko,
  Grigorev, Earl, Dash, daniele, Vyshnya, maykulkarni, Hora, Vera, Mangal, and
  Baranowski}]{ruslan_kuprieiev_2022_7020480}
Ruslan Kuprieiev, skshetry, Dmitry Petrov, Paweł Redzyński, Peter Rowlands,
  Casper da~Costa-Luis, Alexander Schepanovski, Ivan Shcheklein, Batuhan
  Taskaya, Gao, Jorge Orpinel, David de~la Iglesia~Castro, Fábio Santos, Aman
  Sharma, Zhanibek, Dani Hodovic, Dave Berenbaum, Nikita Kodenko, Andrew
  Grigorev, Earl, Nabanita Dash, daniele, George Vyshnya, maykulkarni, Max
  Hora, Vera, Sanidhya Mangal, and Wojciech Baranowski. 2022.
\newblock \href {https://doi.org/10.5281/zenodo.7020480} {Dvc: Data version
  control - git for data \& models}.

\bibitem[{Kłeczek(2020)}]{Kleczek2020}
Dariusz Kłeczek. 2020.
\newblock Polbert: Attacking polish nlp tasks with transformers.
\newblock In \emph{Proceedings of the PolEval 2020 Workshop}. Institute of
  Computer Science, Polish Academy of Sciences.

\bibitem[{Lan et~al.(2020)Lan, Chen, Goodman, Gimpel, Sharma, and
  Soricut}]{Lan2020ALBERT}
Zhenzhong Lan, Mingda Chen, Sebastian Goodman, Kevin Gimpel, Piyush Sharma, and
  Radu Soricut. 2020.
\newblock \href {https://openreview.net/forum?id=H1eA7AEtvS} {Albert: A lite
  bert for self-supervised learning of language representations}.
\newblock In \emph{International Conference on Learning Representations}.

\bibitem[{Le et~al.(2020)Le, Vial, Frej, Segonne, Coavoux, Lecouteux, Allauzen,
  Crabb{\'e}, Besacier, and Schwab}]{le2020flaubert}
Hang Le, Lo{\"\i}c Vial, Jibril Frej, Vincent Segonne, Maximin Coavoux,
  Benjamin Lecouteux, Alexandre Allauzen, Benoit Crabb{\'e}, Laurent Besacier,
  and Didier Schwab. 2020.
\newblock Flaubert: Unsupervised language model pre-training for french.
\newblock In \emph{Proceedings of the 12th Language Resources and Evaluation
  Conference}, pages 2479--2490.

\bibitem[{Li et~al.(2020)Li, Sun, Meng, Liang, Wu, and Li}]{li-etal-2020-dice}
Xiaoya Li, Xiaofei Sun, Yuxian Meng, Junjun Liang, Fei Wu, and Jiwei Li. 2020.
\newblock \href {https://doi.org/10.18653/v1/2020.acl-main.45} {Dice loss for
  data-imbalanced {NLP} tasks}.
\newblock In \emph{Proceedings of the 58th Annual Meeting of the Association
  for Computational Linguistics}, pages 465--476, Online. Association for
  Computational Linguistics.

\bibitem[{Liang et~al.(2020)Liang, Duan, Gong, Wu, Guo, Qi, Gong, Shou, Jiang,
  Cao, Fan, Zhang, Agrawal, Cui, Wei, Bharti, Qiao, Chen, Wu, Liu, Yang,
  Campos, Majumder, and Zhou}]{liang-etal-2020-xglue}
Yaobo Liang, Nan Duan, Yeyun Gong, Ning Wu, Fenfei Guo, Weizhen Qi, Ming Gong,
  Linjun Shou, Daxin Jiang, Guihong Cao, Xiaodong Fan, Ruofei Zhang, Rahul
  Agrawal, Edward Cui, Sining Wei, Taroon Bharti, Ying Qiao, Jiun-Hung Chen,
  Winnie Wu, Shuguang Liu, Fan Yang, Daniel Campos, Rangan Majumder, and Ming
  Zhou. 2020.
\newblock \href {https://doi.org/10.18653/v1/2020.emnlp-main.484} {{XGLUE}: A
  new benchmark datasetfor cross-lingual pre-training, understanding and
  generation}.
\newblock In \emph{Proceedings of the 2020 Conference on Empirical Methods in
  Natural Language Processing (EMNLP)}, pages 6008--6018, Online. Association
  for Computational Linguistics.

\bibitem[{Liu et~al.(2021)Liu, Yan, Gong, Qi, Zhang, Jiao, Chen, Fu, Shou, Gong
  et~al.}]{liu2021glge}
Dayiheng Liu, Yu~Yan, Yeyun Gong, Weizhen Qi, Hang Zhang, Jian Jiao, Weizhu
  Chen, Jie Fu, Linjun Shou, Ming Gong, et~al. 2021.
\newblock Glge: A new general language generation evaluation benchmark.
\newblock In \emph{Findings of the Association for Computational Linguistics:
  ACL-IJCNLP 2021}, pages 408--420.

\bibitem[{McCann et~al.(2018)McCann, Keskar, Xiong, and
  Socher}]{mccann2018natural}
Bryan McCann, Nitish~Shirish Keskar, Caiming Xiong, and Richard Socher. 2018.
\newblock The natural language decathlon: Multitask learning as question
  answering.
\newblock \emph{arXiv preprint arXiv:1806.08730}.

\bibitem[{Mroczkowski et~al.(2021)Mroczkowski, Rybak, Wr{\'o}blewska, and
  Gawlik}]{mroczkowski-etal-2021-herbert}
Robert Mroczkowski, Piotr Rybak, Alina Wr{\'o}blewska, and Ireneusz Gawlik.
  2021.
\newblock \href {https://www.aclweb.org/anthology/2021.bsnlp-1.1} {{H}er{BERT}:
  Efficiently pretrained transformer-based language model for {P}olish}.
\newblock In \emph{Proceedings of the 8th Workshop on Balto-Slavic Natural
  Language Processing}, pages 1--10, Kiyv, Ukraine. Association for
  Computational Linguistics.

\bibitem[{Ogrodniczuk and Kope{\'c}(2014)}]{ogrodniczuk-kopec-2014-polish}
Maciej Ogrodniczuk and Mateusz Kope{\'c}. 2014.
\newblock \href
  {http://www.lrec-conf.org/proceedings/lrec2014/pdf/1211_Paper.pdf} {The
  {P}olish summaries corpus}.
\newblock In \emph{Proceedings of the Ninth International Conference on
  Language Resources and Evaluation ({LREC}'14)}, pages 3712--3715, Reykjavik,
  Iceland. European Language Resources Association (ELRA).

\bibitem[{Park et~al.(2021)Park, Moon, Kim, Cho, Han, Park, Song, Kim, Song, Oh
  et~al.}]{park2021klue}
Sungjoon Park, Jihyung Moon, Sungdong Kim, Won~Ik Cho, Jiyoon Han, Jangwon
  Park, Chisung Song, Junseong Kim, Yongsook Song, Taehwan Oh, et~al. 2021.
\newblock Klue: Korean language understanding evaluation.
\newblock \emph{arXiv preprint arXiv:2105.09680}.

\bibitem[{Peters et~al.(2018)Peters, Neumann, Iyyer, Gardner, Clark, Lee, and
  Zettlemoyer}]{peters-etal-2018-deep}
Matthew~E. Peters, Mark Neumann, Mohit Iyyer, Matt Gardner, Christopher Clark,
  Kenton Lee, and Luke Zettlemoyer. 2018.
\newblock \href {https://doi.org/10.18653/v1/N18-1202} {Deep contextualized
  word representations}.
\newblock In \emph{Proceedings of the 2018 Conference of the North {A}merican
  Chapter of the Association for Computational Linguistics: Human Language
  Technologies, Volume 1 (Long Papers)}, pages 2227--2237, New Orleans,
  Louisiana. Association for Computational Linguistics.

\bibitem[{Petroni et~al.(2021)Petroni, Piktus, Fan, Lewis, Yazdani, De~Cao,
  Thorne, Jernite, Karpukhin, Maillard et~al.}]{petroni2021kilt}
Fabio Petroni, Aleksandra Piktus, Angela Fan, Patrick Lewis, Majid Yazdani,
  Nicola De~Cao, James Thorne, Yacine Jernite, Vladimir Karpukhin, Jean
  Maillard, et~al. 2021.
\newblock Kilt: a benchmark for knowledge intensive language tasks.
\newblock In \emph{Proceedings of the 2021 Conference of the North American
  Chapter of the Association for Computational Linguistics: Human Language
  Technologies}, pages 2523--2544.

\bibitem[{Pineau et~al.(2021)Pineau, Vincent-Lamarre, Sinha, Larivi{\'{e}}re,
  Beygelzimer, D'Alch{\'{e}}-Buc, Fox, and Larochelle}]{Pineau2021}
Joelle Pineau, Philippe Vincent-Lamarre, Koustuv Sinha, Vincent
  Larivi{\'{e}}re, Alina Beygelzimer, Florence D'Alch{\'{e}}-Buc, Emily Fox,
  and Hugo Larochelle. 2021.
\newblock \href {http://arxiv.org/abs/2003.12206} {{Improving reproducibility
  in machine learning research (a report from the neurips 2019 reproducibility
  program)}}.
\newblock \emph{Journal of Machine Learning Research}, 22:1--20.

\bibitem[{Przepiórkowski et~al.(2012)Przepiórkowski, Bańko, Górski, and
  Lewandowska-Tomaszczyk}]{przepiorkowski_narodowy_2012}
Adam Przepiórkowski, Mirosław Bańko, Rafał~L. Górski, and Barbara
  Lewandowska-Tomaszczyk, editors. 2012.
\newblock \emph{Narodowy korpus języka polskiego}.
\newblock Wydawnictwo Naukowe PWN.

\bibitem[{Pęzik et~al.(2022)Pęzik, Krawentek, Karasińska, Wilk, Rybińska,
  Peljak-Łapińska, Cichosz, Deckert, and Adamczyk}]{pezik_lrec_2022}
Piotr Pęzik, Gosia Krawentek, Sylwia Karasińska, Paweł Wilk, Paulina
  Rybińska, Angelika Peljak-Łapińska, Anna Cichosz, Mikołaj Deckert, and
  Michał Adamczyk. 2022.
\newblock Diabiz -- an annotated corpus of polish call center dialogs.
\newblock In \emph{{Language Resources and Evaluation Conference} {2022}}.
  European Language Resources Association (ELRA).

\bibitem[{Rae et~al.(2021)Rae, Borgeaud, Cai, Millican, Hoffmann, Song,
  Aslanides, Henderson, Ring, Young, Rutherford, Hennigan, Menick, Cassirer,
  Powell, Driessche, Hendricks, Rauh, Huang, Glaese, Welbl, Dathathri, Huang,
  Uesato, Mellor, Higgins, Creswell, McAleese, Wu, Elsen, Jayakumar,
  Buchatskaya, Budden, Sutherland, Simonyan, Paganini, Sifre, Martens, Li,
  Kuncoro, Nematzadeh, Gribovskaya, Donato, Lazaridou, Mensch, Lespiau,
  Tsimpoukelli, Grigorev, Fritz, Sottiaux, Pajarskas, Pohlen, Gong, Toyama,
  d'Autume, Li, Terzi, Mikulik, Babuschkin, Clark, Casas, Guy, Jones, Bradbury,
  Johnson, Hechtman, Weidinger, Gabriel, Isaac, Lockhart, Osindero, Rimell,
  Dyer, Vinyals, Ayoub, Stanway, Bennett, Hassabis, Kavukcuoglu, and
  Irving}]{Rae2021}
Jack~W. Rae, Sebastian Borgeaud, Trevor Cai, Katie Millican, Jordan Hoffmann,
  Francis Song, John Aslanides, Sarah Henderson, Roman Ring, Susannah Young,
  Eliza Rutherford, Tom Hennigan, Jacob Menick, Albin Cassirer, Richard Powell,
  George van~den Driessche, Lisa~Anne Hendricks, Maribeth Rauh, Po-Sen Huang,
  Amelia Glaese, Johannes Welbl, Sumanth Dathathri, Saffron Huang, Jonathan
  Uesato, John Mellor, Irina Higgins, Antonia Creswell, Nat McAleese, Amy Wu,
  Erich Elsen, Siddhant Jayakumar, Elena Buchatskaya, David Budden, Esme
  Sutherland, Karen Simonyan, Michela Paganini, Laurent Sifre, Lena Martens,
  Xiang~Lorraine Li, Adhiguna Kuncoro, Aida Nematzadeh, Elena Gribovskaya,
  Domenic Donato, Angeliki Lazaridou, Arthur Mensch, Jean-Baptiste Lespiau,
  Maria Tsimpoukelli, Nikolai Grigorev, Doug Fritz, Thibault Sottiaux, Mantas
  Pajarskas, Toby Pohlen, Zhitao Gong, Daniel Toyama, Cyprien de~Masson
  d'Autume, Yujia Li, Tayfun Terzi, Vladimir Mikulik, Igor Babuschkin, Aidan
  Clark, Diego de~Las Casas, Aurelia Guy, Chris Jones, James Bradbury, Matthew
  Johnson, Blake Hechtman, Laura Weidinger, Iason Gabriel, William Isaac,
  Ed~Lockhart, Simon Osindero, Laura Rimell, Chris Dyer, Oriol Vinyals, Kareem
  Ayoub, Jeff Stanway, Lorrayne Bennett, Demis Hassabis, Koray Kavukcuoglu, and
  Geoffrey Irving. 2021.
\newblock \href {https://doi.org/10.48550/ARXIV.2112.11446} {Scaling language
  models: Methods, analysis \& insights from training gopher}.

\bibitem[{Reimers and Gurevych(2019)}]{reimers-2019-sentence-bert}
Nils Reimers and Iryna Gurevych. 2019.
\newblock \href {http://arxiv.org/abs/1908.10084} {Sentence-bert: Sentence
  embeddings using siamese bert-networks}.
\newblock In \emph{Proceedings of the 2019 Conference on Empirical Methods in
  Natural Language Processing}. Association for Computational Linguistics.

\bibitem[{Rybak et~al.(2020)Rybak, Mroczkowski, Tracz, and Gawlik}]{klej}
Piotr Rybak, Robert Mroczkowski, Janusz Tracz, and Ireneusz Gawlik. 2020.
\newblock Klej: Comprehensive benchmark for polish language understanding.
\newblock In \emph{Proceedings of the 58th Annual Meeting of the Association
  for Computational Linguistics}, pages 1191--1201.

\bibitem[{Safaya et~al.(2022)Safaya, Kurtulu{\c{s}}, Goktogan, and
  Yuret}]{safaya2022mukayese}
Ali Safaya, Emirhan Kurtulu{\c{s}}, Arda Goktogan, and Deniz Yuret. 2022.
\newblock Mukayese: Turkish nlp strikes back.
\newblock In \emph{Findings of the Association for Computational Linguistics:
  ACL 2022}, pages 846--863.

\bibitem[{Seelawi et~al.(2021)Seelawi, Tuffaha, Gzawi, Farhan, Talafha, Badawi,
  Sober, Al-Dweik, Freihat, and Al-Natsheh}]{seelawi2021alue}
Haitham Seelawi, Ibraheem Tuffaha, Mahmoud Gzawi, Wael Farhan, Bashar Talafha,
  Riham Badawi, Zyad Sober, Oday Al-Dweik, Abed~Alhakim Freihat, and Hussein
  Al-Natsheh. 2021.
\newblock Alue: Arabic language understanding evaluation.
\newblock In \emph{Proceedings of the Sixth Arabic Natural Language Processing
  Workshop}, pages 173--184.

\bibitem[{Shavrina et~al.(2020)Shavrina, Fenogenova, Anton, Shevelev, Artemova,
  Malykh, Mikhailov, Tikhonova, Chertok, and
  Evlampiev}]{shavrina-etal-2020-russiansuperglue}
Tatiana Shavrina, Alena Fenogenova, Emelyanov Anton, Denis Shevelev, Ekaterina
  Artemova, Valentin Malykh, Vladislav Mikhailov, Maria Tikhonova, Andrey
  Chertok, and Andrey Evlampiev. 2020.
\newblock \href {https://doi.org/10.18653/v1/2020.emnlp-main.381}
  {{R}ussian{S}uper{GLUE}: A {R}ussian language understanding evaluation
  benchmark}.
\newblock In \emph{Proceedings of the 2020 Conference on Empirical Methods in
  Natural Language Processing (EMNLP)}, pages 4717--4726, Online. Association
  for Computational Linguistics.

\bibitem[{Wang et~al.(2019{\natexlab{a}})Wang, Pruksachatkun, Nangia, Singh,
  Michael, Hill, Levy, and Bowman}]{Wang2019}
Alex Wang, Yada Pruksachatkun, Nikita Nangia, Amanpreet Singh, Julian Michael,
  Felix Hill, Omer Levy, and Samuel Bowman. 2019{\natexlab{a}}.
\newblock \href
  {https://proceedings.neurips.cc/paper/2019/file/4496bf24afe7fab6f046bf4923da8de6-Paper.pdf}
  {Superglue: A stickier benchmark for general-purpose language understanding
  systems}.
\newblock In \emph{Advances in Neural Information Processing Systems},
  volume~32. Curran Associates, Inc.

\bibitem[{Wang et~al.(2019{\natexlab{b}})Wang, Pruksachatkun, Nangia, Singh,
  Michael, Hill, Levy, and Bowman}]{wang2019superglue}
Alex Wang, Yada Pruksachatkun, Nikita Nangia, Amanpreet Singh, Julian Michael,
  Felix Hill, Omer Levy, and Samuel Bowman. 2019{\natexlab{b}}.
\newblock Superglue: A stickier benchmark for general-purpose language
  understanding systems.
\newblock \emph{Advances in neural information processing systems}, 32.

\bibitem[{Wang et~al.(2018)Wang, Singh, Michael, Hill, Levy, and
  Bowman}]{wang2018glue}
Alex Wang, Amanpreet Singh, Julian Michael, Felix Hill, Omer Levy, and Samuel~R
  Bowman. 2018.
\newblock Glue: A multi-task benchmark and analysis platform for natural
  language understanding.
\newblock \emph{arXiv preprint arXiv:1804.07461}.

\bibitem[{Wang et~al.(2022)Wang, Shen, Peng, Zhang, Xiao, Liu, Tang, Chen, Wu,
  and Wang}]{wang2022fine}
Lijie Wang, Yaozong Shen, Shuyuan Peng, Shuai Zhang, Xinyan Xiao, Hao Liu,
  Hongxuan Tang, Ying Chen, Hua Wu, and Haifeng Wang. 2022.
\newblock A fine-grained interpretability evaluation benchmark for neural nlp.
\newblock \emph{arXiv preprint arXiv:2205.11097}.

\bibitem[{Wang et~al.(2020)Wang, Ji, Wang, Wu, Lin, Li, Ke, Xiao, Jiang, Xu,
  and Zhou}]{WANG2020103418}
Qiong Wang, Zongcheng Ji, Jingqi Wang, Stephen Wu, Weiyan Lin, Wenzhen Li,
  Li~Ke, Guohong Xiao, Qing Jiang, Hua Xu, and Yi~Zhou. 2020.
\newblock \href {https://doi.org/https://doi.org/10.1016/j.jbi.2020.103418} {A
  study of entity-linking methods for normalizing chinese diagnosis and
  procedure terms to icd codes}.
\newblock \emph{Journal of Biomedical Informatics}, 105:103418.

\bibitem[{Wilie et~al.(2020)Wilie, Vincentio, Winata, Cahyawijaya, Li, Lim,
  Soleman, Mahendra, Fung, Bahar et~al.}]{wilie2020indonlu}
Bryan Wilie, Karissa Vincentio, Genta~Indra Winata, Samuel Cahyawijaya,
  Xiaohong Li, Zhi~Yuan Lim, Sidik Soleman, Rahmad Mahendra, Pascale Fung,
  Syafri Bahar, et~al. 2020.
\newblock Indonlu: Benchmark and resources for evaluating indonesian natural
  language understanding.
\newblock In \emph{Proceedings of the 1st Conference of the Asia-Pacific
  Chapter of the Association for Computational Linguistics and the 10th
  International Joint Conference on Natural Language Processing}, pages
  843--857.

\bibitem[{Wolf et~al.(2020)Wolf, Debut, Sanh, Chaumond, Delangue, Moi, Cistac,
  Rault, Louf, Funtowicz, Davison, Shleifer, von Platen, Ma, Jernite, Plu, Xu,
  Le~Scao, Gugger, Drame, Lhoest, and Rush}]{wolf-etal-2020-transformers}
Thomas Wolf, Lysandre Debut, Victor Sanh, Julien Chaumond, Clement Delangue,
  Anthony Moi, Pierric Cistac, Tim Rault, Remi Louf, Morgan Funtowicz, Joe
  Davison, Sam Shleifer, Patrick von Platen, Clara Ma, Yacine Jernite, Julien
  Plu, Canwen Xu, Teven Le~Scao, Sylvain Gugger, Mariama Drame, Quentin Lhoest,
  and Alexander Rush. 2020.
\newblock \href {https://doi.org/10.18653/v1/2020.emnlp-demos.6} {Transformers:
  State-of-the-art natural language processing}.
\newblock In \emph{Proceedings of the 2020 Conference on Empirical Methods in
  Natural Language Processing: System Demonstrations}, pages 38--45, Online.
  Association for Computational Linguistics.

\bibitem[{Wr{\'o}blewska and
  Krasnowska-Kiera{\'s}(2017)}]{wroblewska-krasnowska-kieras-2017-polish}
Alina Wr{\'o}blewska and Katarzyna Krasnowska-Kiera{\'s}. 2017.
\newblock \href {https://doi.org/10.18653/v1/P17-1073} {{P}olish evaluation
  dataset for compositional distributional semantics models}.
\newblock In \emph{Proceedings of the 55th Annual Meeting of the Association
  for Computational Linguistics (Volume 1: Long Papers)}, pages 784--792,
  Vancouver, Canada. Association for Computational Linguistics.

\bibitem[{Xu et~al.(2020)Xu, Hu, Zhang, Li, Cao, Li, Xu, Sun, Yu, Yu
  et~al.}]{xu2020clue}
Liang Xu, Hai Hu, Xuanwei Zhang, Lu~Li, Chenjie Cao, Yudong Li, Yechen Xu, Kai
  Sun, Dian Yu, Cong Yu, et~al. 2020.
\newblock Clue: A chinese language understanding evaluation benchmark.
\newblock In \emph{Proceedings of the 28th International Conference on
  Computational Linguistics}, pages 4762--4772.

\bibitem[{Xu et~al.(2021)Xu, Lu, Yuan, Zhang, Xu, Yuan, Wei, Pan, Tian, Qin
  et~al.}]{xu2021fewclue}
Liang Xu, Xiaojing Lu, Chenyang Yuan, Xuanwei Zhang, Huilin Xu, Hu~Yuan, Guoao
  Wei, Xiang Pan, Xin Tian, Libo Qin, et~al. 2021.
\newblock Fewclue: A chinese few-shot learning evaluation benchmark.
\newblock \emph{arXiv preprint arXiv:2107.07498}.

\bibitem[{Yang et~al.(2019)Yang, Dai, Yang, Carbonell, Salakhutdinov, and
  Le}]{NEURIPS2019_dc6a7e65}
Zhilin Yang, Zihang Dai, Yiming Yang, Jaime Carbonell, Russ~R Salakhutdinov,
  and Quoc~V Le. 2019.
\newblock \href
  {https://proceedings.neurips.cc/paper/2019/file/dc6a7e655d7e5840e66733e9ee67cc69-Paper.pdf}
  {Xlnet: Generalized autoregressive pretraining for language understanding}.
\newblock In \emph{Advances in Neural Information Processing Systems},
  volume~32. Curran Associates, Inc.

\bibitem[{Yao et~al.(2021)Yao, Dong, Guan, Cao, Zhang, Xiao, Wang, Qi, Bao, Nie
  et~al.}]{yao2021cuge}
Yuan Yao, Qingxiu Dong, Jian Guan, Boxi Cao, Zhengyan Zhang, Chaojun Xiao,
  Xiaozhi Wang, Fanchao Qi, Junwei Bao, Jinran Nie, et~al. 2021.
\newblock Cuge: A chinese language understanding and generation evaluation
  benchmark.
\newblock \emph{arXiv preprint arXiv:2112.13610}.

\bibitem[{Ye et~al.(2022)Ye, Lin, Li, and Sun}]{ye-etal-2022-packed}
Deming Ye, Yankai Lin, Peng Li, and Maosong Sun. 2022.
\newblock \href {https://doi.org/10.18653/v1/2022.acl-long.337} {Packed
  levitated marker for entity and relation extraction}.
\newblock In \emph{Proceedings of the 60th Annual Meeting of the Association
  for Computational Linguistics (Volume 1: Long Papers)}, pages 4904--4917,
  Dublin, Ireland. Association for Computational Linguistics.

\bibitem[{Zhang et~al.(2021)Zhang, Chen, Bi, Liang, Li, Shang, Yin, Tan, Xu,
  Huang et~al.}]{zhang2021cblue}
Ningyu Zhang, Mosha Chen, Zhen Bi, Xiaozhuan Liang, Lei Li, Xin Shang, Kangping
  Yin, Chuanqi Tan, Jian Xu, Fei Huang, et~al. 2021.
\newblock Cblue: A chinese biomedical language understanding evaluation
  benchmark.
\newblock \emph{arXiv preprint arXiv:2106.08087}.

\end{thebibliography}

\appendix
\appendix

\section{Appendix}

\subsection{Benchmarks}
\begin{table}[ht!]
\centering
\small
\caption{Reviewed NLP benchmarks for languages with at least 10 million speakers. The benchmarks are sorted by the language and name.}
\label{tab:benchmarks}
\begin{tabular}{lrl}
\toprule
                                              Name &  \# of tasks &                   Languages \\
\midrule
                       ALUE \citep{seelawi2021alue} &           6 &                      Arabic \\
                       CBLUE \citep{zhang2021cblue} &           6 &                     Chinese \\
                            CLUE \citep{xu2020clue} &           7 &                     Chinese \\
                           CUGE \citep{yao2021cuge} &          13 &                     Chinese \\
                      FewCLUE \citep{xu2021fewclue} &           7 &                     Chinese \\
                            LOT \citep{guan2022lot} &           3 &                     Chinese \\
                  DecaNLP \citep{mccann2018natural} &           9 &                     English \\
               Dynabench \citep{kiela2021dynabench} &           5 &                     English \\
                           GLGE \citep{liu2021glge} &           4 &                     English \\
                          GLUE \citep{wang2018glue} &           6 &                     English \\
                       KILT \citep{petroni2021kilt} &           5 &                     English \\
               SentEval \citep{conneau2018senteval} &           7 &                     English \\
                SuperGLUE \citep{wang2019superglue} &           5 &                     English \\
                        FLUE \citep{le2020flaubert} &           7 &                      French \\
     IndicNLPSuite \citep{kakwani2020indicnlpsuite} &          10 &                      Indian \\
             IndoNLG \citep{cahyawijaya2021indonlg} &           4 &                  Indonesian \\
                   IndoNLU \citep{wilie2020indonlu} &           7 &                  Indonesian \\
                    IndoLEM \citep{koto2020indolem} &           7 &                  Indonesian \\
                          KLUE \citep{park2021klue} &           8 &                      Korean \\
                       KOBEST \citep{kim2022kobest} &           5 &                      Korean \\
                XGLUE \citep{liang-etal-2020-xglue} &           9 & Multilingual (13 languages) \\
                        GEM \citep{gehrmann2021gem} &           9 & Multilingual (15 languages) \\
                        XTREME \citep{hu2020xtreme} &           6 & Multilingual (17 languages) \\
                          E-KAR \citep{chen2022kar} &           2 &  Multilingual (2 languages) \\
                               \citep{wang2022fine} &           3 &  Multilingual (2 languages) \\
              ParsiNLU \citep{khashabi2021parsinlu} &           6 &                     Persian \\
                             Persian NLP Benchmark \citep{persian-nlp-benchmark} &           9 &                     Persian \\
                                  KLEJ \citep{klej} &           7 &                      Polish \\
                             PLUE \citep{Gomes2020} &           6 &                   Portugese \\
                    LiRO \citep{dumitrescu2021liro} &          10 &                    Romanian \\
            RuMedBench \citep{blinov2022rumedbench} &           4 &                     Russian \\
RussianSuperGLUE \citep{shavrina-etal-2020-russiansuperglue} &           4 &                     Russian \\
                    GLUES \citep{canete2020spanish} &           6 &                     Spanish \\
              SpanishGLUE \citep{canete2020spanish} &           7 &                     Spanish \\
                Mukayese \citep{safaya2022mukayese} &           8 &                     Turkish \\
\bottomrule
\end{tabular}
\end{table}

\newpage
\subsection{Hyperparameter search}
All hyperparameters used to create the first version of benchmark are presented in Table \ref{tab:hyperparameters}.
\ref{tab:hyperparameters}
\begin{table}[ht!]
\centering
\caption{Hyperparameters for finetuning the language models.}
\label{tab:hyperparameters}
\begin{tabular}{lc}
\hline
\textbf{Max. sequence length}             & 512                                              \\ \hline
\textbf{Classifier dropout}               & [0.0, 0.1, 0.2, 0.3, 0.4, 0.5]                   \\ \hline
\textbf{No. finetuned layers}             & [0, 1, 2, 3, 4]                                  \\ \hline
\textbf{Learning rate}                    & [1e-6, 5e-6, 1e-5, 5e-5, 1e-4, 5e-4, 1e-3, 5e-3] \\ \hline
\textbf{Max. no. epochs}                  & [2, 3, 5, 10, 15, 20]                            \\ \hline
\textbf{Batch size}                       & [16, 32, 64]                                     \\ \hline
\textbf{Optimizer}                        & [Adam, AdamW]                                    \\ \hline
\textbf{Weight decay}                     & [1e-4, 1e-3, 1e-2, 1e-1, 0]                      \\ \hline
\textbf{Adam epsilon}                     & [1e-8, 1e-7, 1e-6, 1e-5, 1e-4]                   \\ \hline
\textbf{Use optimizer scheduler}          & [true, false]                                    \\ \hline
\textbf{Optimizer scheduler warmup steps} & [0, 25, 50, 100, 200]                            \\ \hline
\end{tabular}
\end{table}


\subsection{Metrics}
\label{sec:appendix-metrics}
In this section we provide other metrics that were calculated during experimental phase -- tables \ref{tab:accuracy} to \ref{tab:recall_weighted}.

\begin{table}[ht!]
\centering
\small
\caption{
 Accuracy performance of evaluated models on the test subsets. We present values as the mean and standard deviations over 5 model retrains. The mean rank row is the average of a ranking established on the mean of model retrains. Values marked with \textbf{Bold} present the best results for a single dataset. Additionally, we indicate datasets previously appeared in the KLEJ benchmark with \textcolor{red}{*}. \textbf{WIP} denotes the dataset for which we present preliminary results.
}
\label{tab:accuracy}
\begin{tabular}{lrrrrr}
\toprule
 & \makecell{HerBERT\\(base, cased)} & \makecell{HerBERT\\(large, cased)} & \makecell{PolBERT\\(base, cased)} & \makecell{PolBERT\\(base, \\uncased)} & \makecell{XLM-\\RoBERTa\\(paraphrase)} \\
 
\midrule
\makecell[l]{CDSC-E\textcolor{red}{*}} & $\mathBF{94.02 \pm 0.33}$ & $93.92 \pm 0.16$ & $92.30 \pm 0.25$ & $93.48 \pm 0.28$ & $86.58 \pm 0.68$ \\
\makecell[l]{DYK\textcolor{red}{*}} & $\mathBF{90.40 \pm 0.78}$ & $87.66 \pm 0.22$ & $87.60 \pm 0.64$ & $86.82 \pm 0.65$ & $83.79 \pm 0.88$ \\
\makecell[l]{PolEmo 2.0 \\In-Domain\textcolor{red}{*}} & $90.30 \pm 0.28$ & $\mathBF{90.55 \pm 0.47}$ & $87.53 \pm 0.43$ & $87.59 \pm 0.81$ & $85.57 \pm 0.46$ \\
\makecell[l]{PolEmo 2.0 \\Out-Domain\textcolor{red}{*}} & $75.06 \pm 1.86$ & $\mathBF{75.30 \pm 1.69}$ & $69.31 \pm 2.87$ & $69.84 \pm 1.34$ & $57.69 \pm 5.07$ \\
\makecell[l]{PSC\textcolor{red}{*}} & $98.22 \pm 0.20$ & $98.59 \pm 0.57$ & $\mathBF{99.11 \pm 0.11}$ & $99.04 \pm 0.08$ & $73.99 \pm 0.54$ \\
\midrule
\makecell[l]{Abusive \\Clauses} & $87.04 \pm 0.54$ & $\mathBF{88.01 \pm 0.59}$ & $87.49 \pm 0.58$ & $87.13 \pm 0.74$ & $86.42 \pm 0.38$ \\
\makecell[l]{AspectEmo}  & $95.19 \pm 0.07$ & $\mathBF{95.27 \pm 0.24}$ & $94.56 \pm 0.11$ & $94.65 \pm 0.07$ & $92.73 \pm 0.25$ \\
\makecell[l]{KPWr NER} & $97.13 \pm 0.05$ & $\mathBF{97.25 \pm 0.04}$ & $96.86 \pm 0.04$ & $95.90 \pm 0.03$ & $95.74 \pm 0.03$ \\
\makecell[l]{NKJP POS} & $98.88 \pm 0.02$ & $\mathBF{98.98 \pm 0.00}$ & $98.77 \pm 0.02$ & $98.79 \pm 0.02$ & $98.16 \pm 0.02$ \\
\makecell[l]{PolEmo 2.0} & $88.20 \pm 0.50$ & $\mathBF{90.71 \pm 0.40}$ & $87.05 \pm 1.23$ & $87.05 \pm 0.50$ & $85.22 \pm 0.70$ \\
\makecell[l]{Political \\Advertising}  & $96.46 \pm 0.21$ & $\mathBF{96.49 \pm 0.18}$ & $96.12 \pm 0.09$ & $96.38 \pm 0.10$ & $95.71 \pm 0.08$ \\
\makecell[l]{Punctuation \\Restoration} & $93.56 \pm 0.08$ & $\mathBF{94.10 \pm 0.06}$ & $91.89 \pm 0.26$ & $92.38 \pm 0.13$ & $83.71 \pm 0.19$ \\
\makecell[l]{Dialogue Acts\\(WIP)} & $76.50 \pm 0.21$ & $\mathBF{77.08 \pm 0.40}$ & $76.30 \pm 0.25$ & $76.16 \pm 0.24$ & $76.65 \pm 0.75$ \\
\midrule
Mean rank & 2.23 & 1.31 & 3.42 & 3.27 & 4.77 \\
\bottomrule 
\end{tabular}

\end{table}


\begin{table}[ht!]
\centering
\small
\caption{
 Micro F1 performance of evaluated models on the test subsets. We present values as the mean and standard deviations over 5 model retrains. The mean rank row is the average of a ranking established on the mean of model retrains. Values marked with \textbf{Bold} present the best results for a single dataset. Additionally, we indicate datasets previously appeared in the KLEJ benchmark with \textcolor{red}{*}. \textbf{WIP} denotes the dataset for which we present preliminary results.
}
\label{tab:f1_micro}
\begin{tabular}{lrrrrr}
\toprule
 & \makecell{HerBERT\\(base, cased)} & \makecell{HerBERT\\(large, cased)} & \makecell{PolBERT\\(base, cased)} & \makecell{PolBERT\\(base, \\uncased)} & \makecell{XLM-\\RoBERTa\\(paraphrase)} \\
 
\midrule
\makecell[l]{CDSC-E\textcolor{red}{*}} & $\mathBF{94.02 \pm 0.33}$ & $93.92 \pm 0.16$ & $92.30 \pm 0.25$ & $93.48 \pm 0.28$ & $86.58 \pm 0.68$ \\
\makecell[l]{DYK\textcolor{red}{*}} & $\mathBF{90.40 \pm 0.78}$ & $87.66 \pm 0.22$ & $87.60 \pm 0.64$ & $86.82 \pm 0.65$ & $83.79 \pm 0.88$ \\
\makecell[l]{PolEmo 2.0 \\In-Domain\textcolor{red}{*}} & $90.30 \pm 0.28$ & $\mathBF{90.55 \pm 0.47}$ & $87.53 \pm 0.43$ & $87.59 \pm 0.81$ & $85.57 \pm 0.46$ \\
\makecell[l]{PolEmo 2.0 \\Out-Domain\textcolor{red}{*}} & $75.06 \pm 1.86$ & $\mathBF{75.30 \pm 1.69}$ & $69.31 \pm 2.87$ & $69.84 \pm 1.34$ & $57.69 \pm 5.07$ \\
\makecell[l]{PSC\textcolor{red}{*}} & $98.22 \pm 0.20$ & $98.59 \pm 0.57$ & $\mathBF{99.11 \pm 0.11}$ & $99.04 \pm 0.08$ & $73.99 \pm 0.54$ \\
\midrule
\makecell[l]{Abusive \\Clauses} & $87.04 \pm 0.54$ & $\mathBF{88.01 \pm 0.59}$ & $87.49 \pm 0.58$ & $87.13 \pm 0.74$ & $86.42 \pm 0.38$ \\
\makecell[l]{AspectEmo}  & $58.18 \pm 0.32$ & $\mathBF{59.10 \pm 1.32}$ & $50.82 \pm 0.42$ & $51.55 \pm 0.98$ & $34.07 \pm 1.36$ \\
\makecell[l]{KPWr NER} & $\mathBF{76.90 \pm 0.25}$ & $76.58 \pm 0.67$ & $72.59 \pm 0.32$ & $66.49 \pm 0.09$ & $61.57 \pm 0.44$ \\
\makecell[l]{NKJP POS} & $98.88 \pm 0.02$ & $\mathBF{98.98 \pm 0.00}$ & $98.77 \pm 0.02$ & $98.79 \pm 0.02$ & $98.16 \pm 0.02$ \\
\makecell[l]{PolEmo 2.0} & $88.20 \pm 0.50$ & $\mathBF{90.71 \pm 0.40}$ & $87.05 \pm 1.23$ & $87.05 \pm 0.50$ & $85.22 \pm 0.70$ \\
\makecell[l]{Political \\Advertising}  & $68.02 \pm 1.35$ & $67.86 \pm 0.88$ & $65.25 \pm 0.54$ & $\mathBF{68.93 \pm 1.31}$ & $60.94 \pm 0.71$ \\
\makecell[l]{Punctuation \\Restoration} & $73.23 \pm 0.33$ & $\mathBF{75.01 \pm 0.23}$ & $66.49 \pm 0.41$ & $68.56 \pm 0.39$ & $21.72 \pm 0.85$ \\
\makecell[l]{Dialogue Acts\\(WIP)} & $76.50 \pm 0.21$ & $\mathBF{77.08 \pm 0.40}$ & $76.30 \pm 0.25$ & $76.16 \pm 0.24$ & $76.65 \pm 0.75$ \\
\midrule
Mean rank & 2.15 & 1.54 & 3.42 & 3.12 & 4.77 \\
\bottomrule 
\end{tabular}

\end{table}

\begin{table}[ht!]
\centering
\small
\caption{
 Micro Precision performance of evaluated models on the test subsets. We present values as the mean and standard deviations over 5 model retrains. The mean rank row is the average of a ranking established on the mean of model retrains. Values marked with \textbf{Bold} present the best results for a single dataset. Additionally, we indicate datasets previously appeared in the KLEJ benchmark with \textcolor{red}{*}. \textbf{WIP} denotes the dataset for which we present preliminary results.
}
\label{tab:precision_micro}
\begin{tabular}{lrrrrr}
\toprule
 & \makecell{HerBERT\\(base, cased)} & \makecell{HerBERT\\(large, cased)} & \makecell{PolBERT\\(base, cased)} & \makecell{PolBERT\\(base, \\uncased)} & \makecell{XLM-\\RoBERTa\\(paraphrase)} \\
 
\midrule
\makecell[l]{CDSC-E\textcolor{red}{*}} & $\mathBF{94.02 \pm 0.33}$ & $93.92 \pm 0.16$ & $92.30 \pm 0.25$ & $93.48 \pm 0.28$ & $86.58 \pm 0.68$ \\
\makecell[l]{DYK\textcolor{red}{*}} & $\mathBF{90.40 \pm 0.78}$ & $87.66 \pm 0.22$ & $87.60 \pm 0.64$ & $86.82 \pm 0.65$ & $83.79 \pm 0.88$ \\
\makecell[l]{PolEmo 2.0 \\In-Domain\textcolor{red}{*}} & $90.30 \pm 0.28$ & $\mathBF{90.55 \pm 0.47}$ & $87.53 \pm 0.43$ & $87.59 \pm 0.81$ & $85.57 \pm 0.46$ \\
\makecell[l]{PolEmo 2.0 \\Out-Domain\textcolor{red}{*}} & $75.06 \pm 1.86$ & $\mathBF{75.30 \pm 1.69}$ & $69.31 \pm 2.87$ & $69.84 \pm 1.34$ & $57.69 \pm 5.07$ \\
\makecell[l]{PSC\textcolor{red}{*}} & $98.22 \pm 0.20$ & $98.59 \pm 0.57$ & $\mathBF{99.11 \pm 0.11}$ & $99.04 \pm 0.08$ & $73.99 \pm 0.54$ \\
\midrule
\makecell[l]{Abusive \\Clauses} & $87.04 \pm 0.54$ & $\mathBF{88.01 \pm 0.59}$ & $87.49 \pm 0.58$ & $87.13 \pm 0.74$ & $86.42 \pm 0.38$ \\
\makecell[l]{AspectEmo}  & $60.56 \pm 0.99$ & $\mathBF{61.07 \pm 2.81}$ & $55.71 \pm 1.16$ & $56.05 \pm 1.21$ & $40.39 \pm 2.12$ \\
\makecell[l]{KPWr NER} & $\mathBF{75.09 \pm 0.37}$ & $73.85 \pm 0.72$ & $70.30 \pm 0.34$ & $64.48 \pm 0.20$ & $57.28 \pm 0.46$ \\
\makecell[l]{NKJP POS} & $98.88 \pm 0.02$ & $\mathBF{98.98 \pm 0.00}$ & $98.77 \pm 0.02$ & $98.79 \pm 0.02$ & $98.16 \pm 0.02$ \\
\makecell[l]{PolEmo 2.0} & $88.20 \pm 0.50$ & $\mathBF{90.71 \pm 0.40}$ & $87.05 \pm 1.23$ & $87.05 \pm 0.50$ & $85.22 \pm 0.70$ \\
\makecell[l]{Political \\Advertising}  & $64.46 \pm 2.95$ & $64.07 \pm 2.50$ & $66.09 \pm 2.36$ & $\mathBF{68.18 \pm 1.81}$ & $58.85 \pm 2.23$ \\
\makecell[l]{Punctuation \\Restoration} & $74.93 \pm 0.33$ & $\mathBF{77.33 \pm 0.55}$ & $68.50 \pm 2.34$ & $70.64 \pm 1.64$ & $33.43 \pm 1.44$ \\
\makecell[l]{Dialogue Acts\\(WIP)} & $76.50 \pm 0.21$ & $\mathBF{77.08 \pm 0.40}$ & $76.30 \pm 0.25$ & $76.16 \pm 0.24$ & $76.65 \pm 0.75$ \\
\midrule
Mean rank & 2.23 & 1.62 & 3.27 & 3.12 & 4.77 \\
\bottomrule 
\end{tabular}

\end{table}

\begin{table}[ht!]
\centering
\small
\caption{
 Micro Recall performance of evaluated models on the test subsets. We present values as the mean and standard deviations over 5 model retrains. The mean rank row is the average of a ranking established on the mean of model retrains. Values marked with \textbf{Bold} present the best results for a single dataset. Additionally, we indicate datasets previously appeared in the KLEJ benchmark with \textcolor{red}{*}. \textbf{WIP} denotes the dataset for which we present preliminary results.
}
\label{tab:recall_micro}
\begin{tabular}{lrrrrr}
\toprule
 & \makecell{HerBERT\\(base, cased)} & \makecell{HerBERT\\(large, cased)} & \makecell{PolBERT\\(base, cased)} & \makecell{PolBERT\\(base, \\uncased)} & \makecell{XLM-\\RoBERTa\\(paraphrase)} \\
 
\midrule
\makecell[l]{CDSC-E\textcolor{red}{*}} & $\mathBF{94.02 \pm 0.33}$ & $93.92 \pm 0.16$ & $92.30 \pm 0.25$ & $93.48 \pm 0.28$ & $86.58 \pm 0.68$ \\
\makecell[l]{DYK\textcolor{red}{*}} & $\mathBF{90.40 \pm 0.78}$ & $87.66 \pm 0.22$ & $87.60 \pm 0.64$ & $86.82 \pm 0.65$ & $83.79 \pm 0.88$ \\
\makecell[l]{PolEmo 2.0 \\In-Domain\textcolor{red}{*}} & $90.30 \pm 0.28$ & $\mathBF{90.55 \pm 0.47}$ & $87.53 \pm 0.43$ & $87.59 \pm 0.81$ & $85.57 \pm 0.46$ \\
\makecell[l]{PolEmo 2.0 \\Out-Domain\textcolor{red}{*}} & $75.06 \pm 1.86$ & $\mathBF{75.30 \pm 1.69}$ & $69.31 \pm 2.87$ & $69.84 \pm 1.34$ & $57.69 \pm 5.07$ \\
\makecell[l]{PSC\textcolor{red}{*}} & $98.22 \pm 0.20$ & $98.59 \pm 0.57$ & $\mathBF{99.11 \pm 0.11}$ & $99.04 \pm 0.08$ & $73.99 \pm 0.54$ \\
\midrule
\makecell[l]{Abusive \\Clauses} & $87.04 \pm 0.54$ & $\mathBF{88.01 \pm 0.59}$ & $87.49 \pm 0.58$ & $87.13 \pm 0.74$ & $86.42 \pm 0.38$ \\
\makecell[l]{AspectEmo}  & $56.01 \pm 1.09$ & $\mathBF{57.31 \pm 0.95}$ & $46.74 \pm 0.89$ & $47.77 \pm 1.76$ & $29.59 \pm 2.39$ \\
\makecell[l]{KPWr NER} & $78.80 \pm 0.13$ & $\mathBF{79.53 \pm 0.64}$ & $75.05 \pm 0.35$ & $68.63 \pm 0.10$ & $66.55 \pm 0.51$ \\
\makecell[l]{NKJP POS} & $98.88 \pm 0.02$ & $\mathBF{98.98 \pm 0.00}$ & $98.77 \pm 0.02$ & $98.79 \pm 0.02$ & $98.16 \pm 0.02$ \\
\makecell[l]{PolEmo 2.0} & $88.20 \pm 0.50$ & $\mathBF{90.71 \pm 0.40}$ & $87.05 \pm 1.23$ & $87.05 \pm 0.50$ & $85.22 \pm 0.70$ \\
\makecell[l]{Political \\Advertising}  & $72.09 \pm 0.70$ & $\mathBF{72.21 \pm 1.17}$ & $64.53 \pm 1.70$ & $69.70 \pm 1.12$ & $63.30 \pm 1.99$ \\
\makecell[l]{Punctuation \\Restoration} & $71.61 \pm 0.38$ & $\mathBF{72.84 \pm 0.81}$ & $64.70 \pm 1.99$ & $66.65 \pm 1.74$ & $16.10 \pm 0.74$ \\
\makecell[l]{Dialogue Acts\\(WIP)} & $76.50 \pm 0.21$ & $\mathBF{77.08 \pm 0.40}$ & $76.30 \pm 0.25$ & $76.16 \pm 0.24$ & $76.65 \pm 0.75$ \\
\midrule
Mean rank & 2.23 & 1.31 & 3.42 & 3.27 & 4.77 \\
\bottomrule 
\end{tabular}

\end{table}


\begin{table}[ht!]
\centering
\small
\caption{
 Macro F1 performance of evaluated models on the test subsets. We present values as the mean and standard deviations over 5 model retrains. The mean rank row is the average of a ranking established on the mean of model retrains. Values marked with \textbf{Bold} present the best results for a single dataset. Additionally, we indicate datasets previously appeared in the KLEJ benchmark with \textcolor{red}{*}. \textbf{WIP} denotes the dataset for which we present preliminary results.
}
\label{tab:f1_macro}
\begin{tabular}{lrrrrr}
\toprule
 & \makecell{HerBERT\\(base, cased)} & \makecell{HerBERT\\(large, cased)} & \makecell{PolBERT\\(base, cased)} & \makecell{PolBERT\\(base, \\uncased)} & \makecell{XLM-\\RoBERTa\\(paraphrase)} \\
 
\midrule
\makecell[l]{CDSC-E\textcolor{red}{*}} & $\mathBF{90.96 \pm 0.73}$ & $90.48 \pm 0.20$ & $88.95 \pm 0.31$ & $90.62 \pm 0.27$ & $82.62 \pm 0.88$ \\
\makecell[l]{DYK\textcolor{red}{*}} & $\mathBF{82.39 \pm 1.43}$ & $79.58 \pm 0.59$ & $75.87 \pm 0.98$ & $74.41 \pm 1.15$ & $58.93 \pm 7.98$ \\
\makecell[l]{PolEmo 2.0 \\In-Domain\textcolor{red}{*}} & $88.10 \pm 0.36$ & $\mathBF{88.34 \pm 0.63}$ & $85.32 \pm 0.45$ & $85.71 \pm 0.40$ & $83.75 \pm 0.45$ \\
\makecell[l]{PolEmo 2.0 \\Out-Domain\textcolor{red}{*}} & $\mathBF{57.31 \pm 2.93}$ & $57.08 \pm 2.03$ & $54.10 \pm 3.82$ & $54.29 \pm 1.83$ & $45.12 \pm 3.40$ \\
\makecell[l]{PSC\textcolor{red}{*}} & $97.90 \pm 0.24$ & $98.33 \pm 0.69$ & $\mathBF{98.95 \pm 0.13}$ & $98.87 \pm 0.10$ & $58.85 \pm 1.49$ \\
\midrule
\makecell[l]{Abusive \\Clauses} & $85.66 \pm 0.58$ & $\mathBF{86.57 \pm 0.91}$ & $85.93 \pm 0.66$ & $85.74 \pm 0.86$ & $84.32 \pm 0.71$ \\
\makecell[l]{AspectEmo}  & $37.28 \pm 0.71$ & $\mathBF{39.44 \pm 1.74}$ & $30.01 \pm 0.58$ & $31.48 \pm 1.06$ & $18.42 \pm 0.98$ \\
\makecell[l]{KPWr NER} & $\mathBF{54.22 \pm 0.76}$ & $52.68 \pm 1.39$ & $48.01 \pm 0.76$ & $40.21 \pm 0.50$ & $36.13 \pm 0.44$ \\
\makecell[l]{NKJP POS} & $94.59 \pm 0.56$ & $\mathBF{96.14 \pm 0.38}$ & $94.34 \pm 0.61$ & $94.54 \pm 0.19$ & $90.29 \pm 0.51$ \\
\makecell[l]{PolEmo 2.0} & $86.78 \pm 0.79$ & $\mathBF{89.33 \pm 0.49}$ & $85.89 \pm 1.25$ & $85.83 \pm 0.47$ & $84.12 \pm 0.47$ \\
\makecell[l]{Political \\Advertising}  & $61.42 \pm 1.38$ & $62.16 \pm 0.14$ & $58.94 \pm 1.92$ & $\mathBF{62.52 \pm 1.23}$ & $56.68 \pm 0.94$ \\
\makecell[l]{Punctuation \\Restoration} & $45.59 \pm 0.38$ & $\mathBF{46.68 \pm 0.61}$ & $38.89 \pm 0.91$ & $41.31 \pm 0.59$ & $14.33 \pm 1.94$ \\
\makecell[l]{Dialogue Acts\\(WIP)} & $49.54 \pm 0.74$ & $\mathBF{51.11 \pm 0.85}$ & $50.20 \pm 1.32$ & $48.87 \pm 0.90$ & $49.05 \pm 0.39$ \\
\midrule
Mean rank & 2.15 & 1.62 & 3.23 & 3.08 & 4.92 \\
\bottomrule 
\end{tabular}

\end{table}

\begin{table}[ht!]
\centering
\small
\caption{
 Macro Precision performance of evaluated models on the test subsets. We present values as the mean and standard deviations over 5 model retrains. The mean rank row is the average of a ranking established on the mean of model retrains. Values marked with \textbf{Bold} present the best results for a single dataset. Additionally, we indicate datasets previously appeared in the KLEJ benchmark with \textcolor{red}{*}. \textbf{WIP} denotes the dataset for which we present preliminary results.
}
\label{tab:precision_macro}
\begin{tabular}{lrrrrr}
\toprule
 & \makecell{HerBERT\\(base, cased)} & \makecell{HerBERT\\(large, cased)} & \makecell{PolBERT\\(base, cased)} & \makecell{PolBERT\\(base, \\uncased)} & \makecell{XLM-\\RoBERTa\\(paraphrase)} \\
 
\midrule
\makecell[l]{CDSC-E\textcolor{red}{*}} & $92.24 \pm 0.81$ & $\mathBF{92.83 \pm 0.69}$ & $89.91 \pm 1.30$ & $91.20 \pm 0.92$ & $80.18 \pm 1.24$ \\
\makecell[l]{DYK\textcolor{red}{*}} & $\mathBF{83.47 \pm 1.57}$ & $77.85 \pm 0.34$ & $78.90 \pm 1.65$ & $77.30 \pm 1.94$ & $66.03 \pm 13.81$ \\
\makecell[l]{PolEmo 2.0 \\In-Domain\textcolor{red}{*}} & $89.40 \pm 0.55$ & $\mathBF{89.60 \pm 0.95}$ & $85.99 \pm 0.53$ & $85.86 \pm 0.95$ & $83.86 \pm 0.39$ \\
\makecell[l]{PolEmo 2.0 \\Out-Domain\textcolor{red}{*}} & $59.01 \pm 1.95$ & $\mathBF{59.29 \pm 1.80}$ & $56.17 \pm 2.23$ & $56.94 \pm 1.52$ & $52.14 \pm 2.03$ \\
\makecell[l]{PSC\textcolor{red}{*}} & $97.81 \pm 0.23$ & $98.42 \pm 0.34$ & $\mathBF{98.87 \pm 0.11}$ & $98.64 \pm 0.08$ & $74.30 \pm 0.90$ \\
\midrule
\makecell[l]{Abusive \\Clauses} & $84.94 \pm 0.57$ & $\mathBF{86.08 \pm 0.46}$ & $85.54 \pm 0.64$ & $85.18 \pm 0.71$ & $84.99 \pm 0.97$ \\
\makecell[l]{AspectEmo}  & $40.08 \pm 1.12$ & $41.70 \pm 3.99$ & $\mathBF{43.69 \pm 3.19}$ & $37.97 \pm 3.17$ & $23.57 \pm 2.15$ \\
\makecell[l]{KPWr NER} & $\mathBF{55.97 \pm 1.02}$ & $53.07 \pm 1.77$ & $50.45 \pm 1.32$ & $45.00 \pm 1.25$ & $36.64 \pm 0.35$ \\
\makecell[l]{NKJP POS} & $95.91 \pm 0.82$ & $\mathBF{97.23 \pm 0.38}$ & $96.59 \pm 0.73$ & $96.99 \pm 0.23$ & $92.50 \pm 1.17$ \\
\makecell[l]{PolEmo 2.0} & $87.23 \pm 0.43$ & $\mathBF{90.04 \pm 0.61}$ & $86.20 \pm 1.38$ & $86.09 \pm 0.61$ & $84.41 \pm 0.82$ \\
\makecell[l]{Political \\Advertising}  & $58.82 \pm 3.33$ & $59.62 \pm 1.33$ & $60.89 \pm 3.68$ & $\mathBF{62.93 \pm 1.75}$ & $54.68 \pm 2.62$ \\
\makecell[l]{Punctuation \\Restoration} & $49.48 \pm 0.26$ & $\mathBF{53.19 \pm 0.97}$ & $46.22 \pm 2.05$ & $50.29 \pm 6.67$ & $23.85 \pm 1.64$ \\
\makecell[l]{Dialogue Acts\\(WIP)} & $52.83 \pm 0.97$ & $\mathBF{53.61 \pm 1.10}$ & $52.38 \pm 1.06$ & $50.95 \pm 1.10$ & $52.47 \pm 1.65$ \\
\midrule
Mean rank & 2.69 & 1.62 & 2.77 & 3.15 & 4.77 \\
\bottomrule 
\end{tabular}

\end{table}

\begin{table}[ht!]
\centering
\small
\caption{
 Macro Recall performance of evaluated models on the test subsets. We present values as the mean and standard deviations over 5 model retrains. The mean rank row is the average of a ranking established on the mean of model retrains. Values marked with \textbf{Bold} present the best results for a single dataset. Additionally, we indicate datasets previously appeared in the KLEJ benchmark with \textcolor{red}{*}. \textbf{WIP} denotes the dataset for which we present preliminary results.
}
\label{tab:recall_macro}
\begin{tabular}{lrrrrr}
\toprule
 & \makecell{HerBERT\\(base, cased)} & \makecell{HerBERT\\(large, cased)} & \makecell{PolBERT\\(base, cased)} & \makecell{PolBERT\\(base, \\uncased)} & \makecell{XLM-\\RoBERTa\\(paraphrase)} \\
 
\midrule
\makecell[l]{CDSC-E\textcolor{red}{*}} & $89.92 \pm 0.94$ & $88.51 \pm 0.67$ & $88.33 \pm 1.09$ & $\mathBF{90.15 \pm 0.52}$ & $85.78 \pm 1.95$ \\
\makecell[l]{DYK\textcolor{red}{*}} & $81.45 \pm 1.58$ & $\mathBF{81.95 \pm 1.61}$ & $73.77 \pm 0.94$ & $72.52 \pm 1.69$ & $58.11 \pm 5.10$ \\
\makecell[l]{PolEmo 2.0 \\In-Domain\textcolor{red}{*}} & $87.40 \pm 0.33$ & $\mathBF{87.62 \pm 0.86}$ & $84.83 \pm 0.56$ & $85.75 \pm 0.23$ & $83.71 \pm 0.67$ \\
\makecell[l]{PolEmo 2.0 \\Out-Domain\textcolor{red}{*}} & \mathBF{$60.99 \pm 11.77$} & $56.17 \pm 1.53$ & $\mathBF{71.26 \pm 13.42}$ & $66.69 \pm 13.84$ & $42.76 \pm 3.81$ \\
\makecell[l]{PSC\textcolor{red}{*}} & $98.00 \pm 0.33$ & $98.25 \pm 1.02$ & $99.03 \pm 0.15$ & $\mathBF{99.10 \pm 0.15}$ & $59.35 \pm 1.01$ \\
\midrule
\makecell[l]{Abusive \\Clauses} & $86.91 \pm 1.09$ & $\mathBF{87.37 \pm 1.73}$ & $86.51 \pm 1.04$ & $87.01 \pm 1.81$ & $84.06 \pm 1.76$ \\
\makecell[l]{AspectEmo}  & $37.36 \pm 0.72$ & $\mathBF{39.80 \pm 1.30}$ & $27.54 \pm 0.69$ & $29.80 \pm 0.99$ & $16.06 \pm 1.17$ \\
\makecell[l]{KPWr NER} & $\mathBF{55.74 \pm 0.71}$ & $55.52 \pm 1.47$ & $49.03 \pm 0.55$ & $40.81 \pm 0.27$ & $38.60 \pm 0.47$ \\
\makecell[l]{NKJP POS} & $93.95 \pm 0.64$ & $\mathBF{95.39 \pm 0.45}$ & $93.18 \pm 0.50$ & $93.28 \pm 0.12$ & $88.86 \pm 0.33$ \\
\makecell[l]{PolEmo 2.0} & $86.50 \pm 1.00$ & $\mathBF{88.94 \pm 0.39}$ & $85.68 \pm 1.13$ & $85.63 \pm 0.40$ & $83.95 \pm 0.36$ \\
\makecell[l]{Political \\Advertising}  & $65.04 \pm 1.05$ & $\mathBF{65.60 \pm 1.59}$ & $57.70 \pm 1.62$ & $62.72 \pm 1.13$ & $59.55 \pm 1.23$ \\
\makecell[l]{Punctuation \\Restoration} & $43.16 \pm 0.36$ & $\mathBF{43.56 \pm 1.06}$ & $35.50 \pm 1.27$ & $38.11 \pm 0.98$ & $10.71 \pm 1.90$ \\
\makecell[l]{Dialogue Acts\\(WIP)} & $49.54 \pm 1.16$ & $\mathBF{51.79 \pm 0.71}$ & $50.91 \pm 1.45$ & $49.58 \pm 0.81$ & $49.36 \pm 0.84$ \\
\midrule
Mean rank & 2.38 & 1.62 & 3.31 & 2.77 & 4.92 \\
\bottomrule 
\end{tabular}

\end{table}


\begin{table}[ht!]
\centering
\small
\caption{
 Weighted F1 performance of evaluated models on the test subsets. We present values as the mean and standard deviations over 5 model retrains. The mean rank row is the average of a ranking established on the mean of model retrains. Values marked with \textbf{Bold} present the best results for a single dataset. Additionally, we indicate datasets previously appeared in the KLEJ benchmark with \textcolor{red}{*}. \textbf{WIP} denotes the dataset for which we present preliminary results.
}
\label{tab:f1_weighted}
\begin{tabular}{lrrrrr}
\toprule
 & \makecell{HerBERT\\(base, cased)} & \makecell{HerBERT\\(large, cased)} & \makecell{PolBERT\\(base, cased)} & \makecell{PolBERT\\(base, \\uncased)} & \makecell{XLM-\\RoBERTa\\(paraphrase)} \\
 
\midrule
\makecell[l]{CDSC-E\textcolor{red}{*}} & $\mathBF{93.93 \pm 0.32}$ & $93.80 \pm 0.18$ & $92.16 \pm 0.24$ & $93.42 \pm 0.25$ & $86.83 \pm 0.54$ \\
\makecell[l]{DYK\textcolor{red}{*}} & $\mathBF{90.25 \pm 0.79}$ & $88.08 \pm 0.21$ & $87.00 \pm 0.58$ & $86.20 \pm 0.56$ & $80.06 \pm 2.89$ \\
\makecell[l]{PolEmo 2.0 \\In-Domain\textcolor{red}{*}} & $89.90 \pm 0.28$ & $\mathBF{90.20 \pm 0.51}$ & $87.33 \pm 0.45$ & $87.64 \pm 0.53$ & $85.69 \pm 0.40$ \\
\makecell[l]{PolEmo 2.0 \\Out-Domain\textcolor{red}{*}} & $75.61 \pm 2.84$ & $\mathBF{76.10 \pm 2.67}$ & $69.68 \pm 3.43$ & $71.05 \pm 1.20$ & $60.33 \pm 4.55$ \\
\makecell[l]{PSC\textcolor{red}{*}} & $98.22 \pm 0.20$ & $98.59 \pm 0.58$ & $\mathBF{99.11 \pm 0.11}$ & $99.04 \pm 0.08$ & $68.62 \pm 1.00$ \\
\midrule
\makecell[l]{Abusive \\Clauses} & $87.22 \pm 0.51$ & $\mathBF{88.11 \pm 0.69}$ & $87.57 \pm 0.57$ & $87.29 \pm 0.73$ & $86.32 \pm 0.46$ \\
\makecell[l]{AspectEmo}  & $58.27 \pm 0.40$ & $\mathBF{59.16 \pm 1.02}$ & $50.38 \pm 0.39$ & $51.50 \pm 0.95$ & $33.45 \pm 1.42$ \\
\makecell[l]{KPWr NER} & $\mathBF{77.09 \pm 0.24}$ & $77.02 \pm 0.69$ & $72.47 \pm 0.27$ & $65.97 \pm 0.17$ & $61.84 \pm 0.47$ \\
\makecell[l]{NKJP POS} & $98.88 \pm 0.02$ & $\mathBF{98.98 \pm 0.00}$ & $98.76 \pm 0.02$ & $98.79 \pm 0.02$ & $98.16 \pm 0.02$ \\
\makecell[l]{PolEmo 2.0} & $88.02 \pm 0.61$ & $\mathBF{90.44 \pm 0.41}$ & $86.99 \pm 1.18$ & $87.08 \pm 0.49$ & $85.17 \pm 0.53$ \\
\makecell[l]{Political \\Advertising}  & $68.10 \pm 1.34$ & $68.10 \pm 0.83$ & $65.28 \pm 0.56$ & $\mathBF{68.89 \pm 1.33}$ & $60.97 \pm 0.61$ \\
\makecell[l]{Punctuation \\Restoration} & $72.41 \pm 0.30$ & $\mathBF{73.77 \pm 0.26}$ & $65.29 \pm 0.48$ & $67.29 \pm 0.38$ & $21.44 \pm 0.78$ \\
\makecell[l]{Dialogue Acts\\(WIP)} & $75.57 \pm 0.30$ & $\mathBF{76.03 \pm 0.18}$ & $75.43 \pm 0.25$ & $75.21 \pm 0.21$ & $75.59 \pm 0.59$ \\
\midrule
Mean rank & 2.19 & 1.5 & 3.46 & 3.08 & 4.77 \\
\bottomrule 
\end{tabular}

\end{table}

\begin{table}[ht!]
\centering
\small
\caption{
 Weighted Precision performance of evaluated models on the test subsets. We present values as the mean and standard deviations over 5 model retrains. The mean rank row is the average of a ranking established on the mean of model retrains. Values marked with \textbf{Bold} present the best results for a single dataset. Additionally, we indicate datasets previously appeared in the KLEJ benchmark with \textcolor{red}{*}. \textbf{WIP} denotes the dataset for which we present preliminary results.
}
\label{tab:precision_weighted}
\begin{tabular}{lrrrrr}
\toprule
 & \makecell{HerBERT\\(base, cased)} & \makecell{HerBERT\\(large, cased)} & \makecell{PolBERT\\(base, cased)} & \makecell{PolBERT\\(base, \\uncased)} & \makecell{XLM-\\RoBERTa\\(paraphrase)} \\
 
\midrule
\makecell[l]{CDSC-E\textcolor{red}{*}} & $\mathBF{93.98 \pm 0.36}$ & $93.90 \pm 0.16$ & $92.21 \pm 0.24$ & $93.43 \pm 0.29$ & $87.47 \pm 0.67$ \\
\makecell[l]{DYK\textcolor{red}{*}} & $\mathBF{90.16 \pm 0.81}$ & $88.78 \pm 0.50$ & $86.78 \pm 0.64$ & $85.98 \pm 0.64$ & $78.95 \pm 5.62$ \\
\makecell[l]{PolEmo 2.0 \\In-Domain\textcolor{red}{*}} & $89.90 \pm 0.31$ & $\mathBF{90.22 \pm 0.57}$ & $87.27 \pm 0.48$ & $87.84 \pm 0.32$ & $85.86 \pm 0.37$ \\
\makecell[l]{PolEmo 2.0 \\Out-Domain\textcolor{red}{*}} & $77.90 \pm 2.30$ & $\mathBF{78.62 \pm 2.63}$ & $72.96 \pm 2.41$ & $74.62 \pm 1.55$ & $69.00 \pm 2.77$ \\
\makecell[l]{PSC\textcolor{red}{*}} & $98.23 \pm 0.20$ & $98.60 \pm 0.56$ & $\mathBF{99.11 \pm 0.11}$ & $99.05 \pm 0.09$ & $74.15 \pm 0.69$ \\
\midrule
\makecell[l]{Abusive \\Clauses} & $87.83 \pm 0.67$ & $\mathBF{88.46 \pm 0.93}$ & $87.81 \pm 0.55$ & $88.05 \pm 1.07$ & $86.55 \pm 0.42$ \\
\makecell[l]{AspectEmo}  & $61.79 \pm 1.23$ & $\mathBF{62.07 \pm 2.14}$ & $56.96 \pm 1.27$ & $57.36 \pm 1.69$ & $39.70 \pm 1.43$ \\
\makecell[l]{KPWr NER} & $\mathBF{76.43 \pm 0.35}$ & $75.60 \pm 0.74$ & $71.13 \pm 0.25$ & $65.41 \pm 0.30$ & $58.98 \pm 0.43$ \\
\makecell[l]{NKJP POS} & $98.88 \pm 0.02$ & $\mathBF{98.99 \pm 0.00}$ & $98.77 \pm 0.02$ & $98.80 \pm 0.02$ & $98.16 \pm 0.02$ \\
\makecell[l]{PolEmo 2.0} & $87.98 \pm 0.60$ & $\mathBF{90.41 \pm 0.43}$ & $87.00 \pm 1.11$ & $87.18 \pm 0.46$ & $85.22 \pm 0.50$ \\
\makecell[l]{Political \\Advertising}  & $64.97 \pm 3.06$ & $64.93 \pm 2.40$ & $66.61 \pm 2.34$ & $\mathBF{68.50 \pm 1.87}$ & $59.17 \pm 2.21$ \\
\makecell[l]{Punctuation \\Restoration} & $73.94 \pm 0.27$ & $\mathBF{76.80 \pm 0.49}$ & $67.80 \pm 1.84$ & $69.84 \pm 1.76$ & $32.96 \pm 1.15$ \\
\makecell[l]{Dialogue Acts\\(WIP)} & $75.96 \pm 0.47$ & $\mathBF{76.30 \pm 0.34}$ & $75.75 \pm 0.55$ & $75.20 \pm 0.27$ & $75.99 \pm 0.86$ \\
\midrule
Mean rank & 2.15 & 1.62 & 3.46 & 3.0 & 4.77 \\
\bottomrule 
\end{tabular}

\end{table}

\begin{table}[ht!]
\centering
\small
\caption{
 Weighted Recall performance of evaluated models on the test subsets. We present values as the mean and standard deviations over 5 model retrains. The mean rank row is the average of a ranking established on the mean of model retrains. Values marked with \textbf{Bold} present the best results for a single dataset. Additionally, we indicate datasets previously appeared in the KLEJ benchmark with \textcolor{red}{*}. \textbf{WIP} denotes the dataset for which we present preliminary results.
}
\label{tab:recall_weighted}
\begin{tabular}{lrrrrr}
\toprule
 & \makecell{HerBERT\\(base, cased)} & \makecell{HerBERT\\(large, cased)} & \makecell{PolBERT\\(base, cased)} & \makecell{PolBERT\\(base, \\uncased)} & \makecell{XLM-\\RoBERTa\\(paraphrase)} \\
 
\midrule
\makecell[l]{CDSC-E\textcolor{red}{*}} & $\mathBF{94.02 \pm 0.33}$ & $93.92 \pm 0.16$ & $92.30 \pm 0.25$ & $93.48 \pm 0.28$ & $86.58 \pm 0.68$ \\
\makecell[l]{DYK\textcolor{red}{*}} & $\mathBF{90.40 \pm 0.78}$ & $87.66 \pm 0.22$ & $87.60 \pm 0.64$ & $86.82 \pm 0.65$ & $83.79 \pm 0.88$ \\
\makecell[l]{PolEmo 2.0 \\In-Domain\textcolor{red}{*}} & $90.30 \pm 0.28$ & $\mathBF{90.55 \pm 0.47}$ & $87.53 \pm 0.43$ & $87.59 \pm 0.81$ & $85.57 \pm 0.46$ \\
\makecell[l]{PolEmo 2.0 \\Out-Domain\textcolor{red}{*}} & $75.06 \pm 1.86$ & $\mathBF{75.30 \pm 1.69}$ & $69.31 \pm 2.87$ & $69.84 \pm 1.34$ & $57.69 \pm 5.07$ \\
\makecell[l]{PSC\textcolor{red}{*}} & $98.22 \pm 0.20$ & $98.59 \pm 0.57$ & $\mathBF{99.11 \pm 0.11}$ & $99.04 \pm 0.08$ & $73.99 \pm 0.54$ \\
\midrule
\makecell[l]{Abusive \\Clauses} & $87.04 \pm 0.54$ & $\mathBF{88.01 \pm 0.59}$ & $87.49 \pm 0.58$ & $87.13 \pm 0.74$ & $86.42 \pm 0.38$ \\
\makecell[l]{AspectEmo}  & $56.01 \pm 1.09$ & $\mathBF{57.31 \pm 0.95}$ & $46.74 \pm 0.89$ & $47.77 \pm 1.76$ & $29.59 \pm 2.39$ \\
\makecell[l]{KPWr NER} & $78.80 \pm 0.13$ & $\mathBF{79.53 \pm 0.64}$ & $75.05 \pm 0.35$ & $68.63 \pm 0.10$ & $66.55 \pm 0.51$ \\
\makecell[l]{NKJP POS} & $98.88 \pm 0.02$ & $\mathBF{98.98 \pm 0.00}$ & $98.77 \pm 0.02$ & $98.79 \pm 0.02$ & $98.16 \pm 0.02$ \\
\makecell[l]{PolEmo 2.0} & $88.20 \pm 0.50$ & $\mathBF{90.71 \pm 0.40}$ & $87.05 \pm 1.23$ & $87.05 \pm 0.50$ & $85.22 \pm 0.70$ \\
\makecell[l]{Political \\Advertising}  & $72.09 \pm 0.70$ & $\mathBF{72.21 \pm 1.17}$ & $64.53 \pm 1.70$ & $69.70 \pm 1.12$ & $63.30 \pm 1.99$ \\
\makecell[l]{Punctuation \\Restoration} & $71.61 \pm 0.38$ & $\mathBF{72.84 \pm 0.81}$ & $64.70 \pm 1.99$ & $66.65 \pm 1.74$ & $16.10 \pm 0.74$ \\
\makecell[l]{Dialogue Acts\\(WIP)} & $76.50 \pm 0.21$ & $\mathBF{77.08 \pm 0.40}$ & $76.30 \pm 0.25$ & $76.16 \pm 0.24$ & $76.65 \pm 0.75$ \\
\midrule
Mean rank & 2.23 & 1.31 & 3.42 & 3.27 & 4.77 \\
\bottomrule 
\end{tabular}

\end{table}


\end{document}